\documentclass[10pt,twocolumn,letterpaper]{article}

\usepackage[pagenumbers]{iccv} 

\usepackage[accsupp]{axessibility}  

\usepackage{tikz}           
\usepackage{pgfplots,pgfplotstable} 
\usepackage{xcolor}         

\definecolor{iccvblue}{rgb}{0.21,0.49,0.74}
\usepackage[pagebackref,breaklinks,colorlinks,allcolors=iccvblue]{hyperref}

\def\paperID{14827} 
\def\confName{ICCV}
\def\confYear{2025}

\title{Image-Guided Shape-from-Template Using Mesh Inextensibility Constraints}

\author{Thuy Tran \qquad Ruochen Chen \qquad Shaifali Parashar \vspace{.15cm}\\
CNRS, École Centrale de Lyon,
INSA Lyon, Université Claude Bernard Lyon 1, LIRIS, UMR5205 \\
{\tt\small \{dinh-vinh-thuy.tran, ruochen.chen, shaifali.parashar\}@liris.cnrs.fr}
}

\begin{document}
\maketitle
\newcommand{\drawDefault}{%
(0, 3.359450101852417)
(1, 3.4212656021118164)
(2, 3.4019367694854736)
(3, 2.9442965984344482)
(4, 3.26605224609375)
(5, 3.119257688522339)
(6, 3.087919235229492)
(7, 3.6031510829925537)
(8, 3.287980794906616)
(9, 3.1907410621643066)
(10, 2.9637765884399414)
(11, 3.0549137592315674)
(12, 3.268265724182129)
(13, 3.5708751678466797)
(14, 3.645174503326416)
(15, 3.290022134780884)
(16, 3.258082628250122)
(17, 3.435577630996704)
(18, 3.419078826904297)
(19, 3.4210562705993652)
(20, 3.4845621585845947)
(21, 3.341963768005371)
(22, 3.140979766845703)
(23, 3.0394811630249023)
(24, 3.252882957458496)
(25, 4.507602691650391)
(26, 4.674405097961426)
(27, 4.967648029327393)
(28, 5.753402233123779)
(29, 6.1287384033203125)
(30, 4.810593128204346)
(31, 4.2301249504089355)
(32, 5.180417537689209)
(33, 3.95287823677063)
(34, 3.4053571224212646)
(35, 2.966266632080078)
(36, 3.151819944381714)
(37, 2.9469187259674072)
(38, 3.0915586948394775)
(39, 2.967633008956909)
(40, 3.0344576835632324)
(41, 2.9393675327301025)
(42, 2.9214954376220703)
(43, 3.0686795711517334)
(44, 2.891338586807251)
(45, 3.034735918045044)
(46, 3.047732353210449)
(47, 3.0938186645507812)
(48, 2.977522850036621)
(49, 2.785687208175659)
(50, 2.6473939418792725)
(51, 2.5856986045837402)
(52, 2.385561466217041)
(53, 2.2737784385681152)
(54, 2.0288000106811523)
(55, 1.8959137201309204)
(56, 3.4553980827331543)
(57, 1.481851577758789)
(58, 0.8297741413116455)
(59, 1.205112099647522)
(60, 1.1797444820404053)
(61, 0.7965299487113953)
(62, 2.7041268348693848)
(63, 1.7514768838882446)
(64, 1.21575927734375)
(65, 1.6892825365066528)
(66, 1.7259347438812256)
(67, 1.3953715562820435)
(68, 1.46957266330719)
(69, 0.8781396150588989)
(70, 3.980682373046875)
(71, 3.7523865699768066)
(72, 2.036660671234131)
(73, 2.499173641204834)
(74, 3.414442539215088)
(75, 4.6905083656311035)
(76, 5.729549884796143)
(77, 3.2792680263519287)
(78, 4.68652868270874)
(79, 2.93361759185791)
(80, 2.056281805038452)
(81, 2.2550888061523438)
(82, 2.3470654487609863)
(83, 2.4929981231689453)
(84, 2.9935905933380127)
(85, 2.617466688156128)
(86, 3.3357717990875244)
(87, 2.95870041847229)
(88, 3.0009541511535645)
(89, 3.5803091526031494)
(90, 3.2785375118255615)
(91, 3.0950546264648438)
(92, 4.627209663391113)
(93, 3.4381701946258545)
(94, 3.6480276584625244)
(95, 3.561000347137451)
(96, 3.671518325805664)
(97, 3.6609086990356445)
(98, 3.6924750804901123)
(99, 4.363518238067627)
(100, 3.9448440074920654)
(101, 3.8025360107421875)
(102, 4.397543907165527)
(103, 5.549528121948242)
(104, 5.111042499542236)
(105, 4.196010112762451)
(106, 4.00147819519043)
(107, 4.429396152496338)
(108, 4.677762031555176)
(109, 5.7884368896484375)
(110, 6.278349876403809)
(111, 5.299497127532959)
(112, 5.166465759277344)
(113, 5.303836822509766)
(114, 7.435054302215576)
(115, 16.897104263305664)
(116, 2.260401725769043)
(117, 1.561222791671753)
(118, 2.5209312438964844)
(119, 3.520895481109619)
(120, 2.7665951251983643)
(121, 3.7525532245635986)
(122, 3.1196022033691406)
(123, 3.369511127471924)
(124, 2.993035316467285)
(125, 3.3713297843933105)
(126, 8.318131446838379)
(127, 3.8383231163024902)
(128, 3.2263996601104736)
(129, 3.4343957901000977)
(130, 3.645328998565674)
(131, 16.029434204101562)
(132, 4.878126621246338)
(133, 3.2358152866363525)
(134, 3.6725127696990967)
(135, 5.328373908996582)
(136, 20.26536750793457)
(137, 12.759010314941406)
(138, 6.630215644836426)
(139, 4.308798789978027)
(140, 2.582245111465454)
(141, 3.505772113800049)
(142, 1.8209092617034912)
(143, 1.8831589221954346)
(144, 7.661078453063965)
(145, 4.142118453979492)
(146, 1.4636621475219727)
(147, 1.3177677392959595)
(148, 1.5383074283599854)
(149, 1.7175006866455078)
(150, 1.7142552137374878)
(151, 1.9285426139831543)
(152, 20.52507972717285)
(153, 3.382403612136841)
(154, 2.8156754970550537)
(155, 3.1844069957733154)
(156, 3.295214891433716)
(157, 3.171976089477539)
(158, 3.241574764251709)
(159, 3.3592233657836914)
(160, 3.880805015563965)
(161, 3.317375421524048)
(162, 4.3921732902526855)
(163, 4.418516159057617)
(164, 3.7365567684173584)
(165, 3.990690231323242)
(166, 6.136383533477783)
(167, 6.976240158081055)
(168, 7.378639221191406)
(169, 7.8651580810546875)
(170, 9.403382301330566)
(171, 9.518104553222656)
(172, 7.459652900695801)
(173, 6.854042053222656)
(174, 6.022702693939209)
(175, 5.288254261016846)
(176, 3.944851875305176)
(177, 3.6949679851531982)
(178, 3.61672043800354)
(179, 3.628101110458374)
(180, 6.712699890136719)
(181, 5.048187732696533)
(182, 4.936291217803955)
(183, 4.7057647705078125)
(184, 4.8840107917785645)
(185, 7.125514507293701)
(186, 2.472933292388916)
(187, 1.9917421340942383)
(188, 2.5540871620178223)
(189, 2.9066967964172363)
(190, 3.1077992916107178)
(191, 3.9013051986694336)
(192, 5.121196269989014)
}

\newcommand{\drawMoreIterations}{%
(0, 3.359450101852417)
(1, 3.421077013015747)
(2, 3.3982484340667725)
(3, 2.9190051555633545)
(4, 3.2405283451080322)
(5, 3.133943796157837)
(6, 3.0958449840545654)
(7, 3.609248638153076)
(8, 3.2168776988983154)
(9, 3.1233720779418945)
(10, 2.890916585922241)
(11, 3.0459446907043457)
(12, 3.2728617191314697)
(13, 3.578275680541992)
(14, 3.6543424129486084)
(15, 3.3115105628967285)
(16, 3.2846717834472656)
(17, 3.4400956630706787)
(18, 3.3919804096221924)
(19, 3.419125556945801)
(20, 3.469006299972534)
(21, 3.3406612873077393)
(22, 3.1537673473358154)
(23, 3.06669020652771)
(24, 3.2283403873443604)
(25, 4.448614120483398)
(26, 4.585952281951904)
(27, 4.7821221351623535)
(28, 5.320163726806641)
(29, 5.914734840393066)
(30, 4.545290946960449)
(31, 3.707287549972534)
(32, 4.739428520202637)
(33, 3.484628200531006)
(34, 2.9803290367126465)
(35, 2.8413290977478027)
(36, 2.614217519760132)
(37, 2.4881811141967773)
(38, 2.8426928520202637)
(39, 2.88238263130188)
(40, 2.9128270149230957)
(41, 2.7641053199768066)
(42, 2.8883886337280273)
(43, 2.8601949214935303)
(44, 2.7278010845184326)
(45, 2.718047857284546)
(46, 2.788135051727295)
(47, 2.870183229446411)
(48, 2.519700288772583)
(49, 2.4585297107696533)
(50, 2.713754415512085)
(51, 2.3670196533203125)
(52, 2.1921746730804443)
(53, 2.129460096359253)
(54, 1.9472101926803589)
(55, 1.9093806743621826)
(56, 3.1730806827545166)
(57, 1.3819468021392822)
(58, 0.7077112197875977)
(59, 1.1300721168518066)
(60, 1.0997552871704102)
(61, 0.8571358323097229)
(62, 2.6510579586029053)
(63, 1.7606213092803955)
(64, 1.2610607147216797)
(65, 1.7266266345977783)
(66, 1.7279376983642578)
(67, 1.4347385168075562)
(68, 1.4368571043014526)
(69, 0.9555324912071228)
(70, 3.93673038482666)
(71, 3.9611001014709473)
(72, 2.2707290649414062)
(73, 2.451840400695801)
(74, 3.4433929920196533)
(75, 4.631412029266357)
(76, 5.317756175994873)
(77, 3.107185125350952)
(78, 4.601255893707275)
(79, 2.3705859184265137)
(80, 2.0242974758148193)
(81, 2.3674347400665283)
(82, 2.307277202606201)
(83, 2.4998326301574707)
(84, 3.019050121307373)
(85, 2.6740736961364746)
(86, 3.356794834136963)
(87, 2.9215939044952393)
(88, 2.9872710704803467)
(89, 3.5080597400665283)
(90, 3.290844202041626)
(91, 3.100198745727539)
(92, 4.645061492919922)
(93, 3.5192854404449463)
(94, 3.7853565216064453)
(95, 3.6611106395721436)
(96, 3.844299077987671)
(97, 3.6902201175689697)
(98, 3.7508671283721924)
(99, 4.361557483673096)
(100, 3.9809398651123047)
(101, 3.9895052909851074)
(102, 4.44391393661499)
(103, 5.681004047393799)
(104, 5.19986629486084)
(105, 4.406341552734375)
(106, 4.215152740478516)
(107, 4.676060676574707)
(108, 4.744421482086182)
(109, 5.7466325759887695)
(110, 6.3129425048828125)
(111, 5.390576362609863)
(112, 5.028404235839844)
(113, 5.151905059814453)
(114, 7.295974254608154)
(115, 17.230987548828125)
(116, 2.227193593978882)
(117, 1.7435665130615234)
(118, 2.7493035793304443)
(119, 3.6592092514038086)
(120, 3.192107915878296)
(121, 4.338179111480713)
(122, 3.286550521850586)
(123, 3.6706550121307373)
(124, 3.361969232559204)
(125, 3.275371789932251)
(126, 7.916676044464111)
(127, 3.4139201641082764)
(128, 2.794926881790161)
(129, 3.247699022293091)
(130, 3.58644437789917)
(131, 15.294650077819824)
(132, 3.472026824951172)
(133, 3.3618075847625732)
(134, 3.718783378601074)
(135, 5.370397090911865)
(136, 19.677200317382812)
(137, 10.920305252075195)
(138, 6.804287910461426)
(139, 3.7969400882720947)
(140, 3.040074110031128)
(141, 3.5320446491241455)
(142, 1.8515427112579346)
(143, 1.8342134952545166)
(144, 7.680769443511963)
(145, 3.9535810947418213)
(146, 1.4599007368087769)
(147, 1.3495405912399292)
(148, 1.5430502891540527)
(149, 1.689536452293396)
(150, 1.7438135147094727)
(151, 1.9379867315292358)
(152, 20.68931007385254)
(153, 3.6309027671813965)
(154, 3.0393142700195312)
(155, 3.0839109420776367)
(156, 3.104440212249756)
(157, 3.2598953247070312)
(158, 3.4440841674804688)
(159, 3.812216281890869)
(160, 3.951570510864258)
(161, 3.362797498703003)
(162, 4.363216876983643)
(163, 4.535543918609619)
(164, 3.9283454418182373)
(165, 4.160345077514648)
(166, 6.599997520446777)
(167, 7.266610145568848)
(168, 8.80831241607666)
(169, 8.206123352050781)
(170, 9.3034029006958)
(171, 9.855770111083984)
(172, 8.649372100830078)
(173, 7.440787315368652)
(174, 6.433746814727783)
(175, 6.1685471534729)
(176, 5.445987701416016)
(177, 3.65498423576355)
(178, 3.4935994148254395)
(179, 3.4623935222625732)
(180, 6.386562824249268)
(181, 4.746792793273926)
(182, 4.45873498916626)
(183, 4.2916579246521)
(184, 4.661316394805908)
(185, 6.975351333618164)
(186, 2.492818832397461)
(187, 2.0069797039031982)
(188, 2.4564743041992188)
(189, 2.898179531097412)
(190, 3.0023000240325928)
(191, 3.9004452228546143)
(192, 5.107312202453613)
}

\newcommand{\drawWindowWise}{%
(0, 3.359450101852417)
(1, 3.4161574840545654)
(2, 3.442648410797119)
(3, 2.9946413040161133)
(4, 3.330120086669922)
(5, 3.1002204418182373)
(6, 2.9609427452087402)
(7, 3.2547290325164795)
(8, 3.0894789695739746)
(9, 2.9707789421081543)
(10, 2.9653522968292236)
(11, 3.0381252765655518)
(12, 3.1275975704193115)
(13, 3.263062000274658)
(14, 3.2037158012390137)
(15, 3.2133376598358154)
(16, 3.2044217586517334)
(17, 3.3204526901245117)
(18, 3.2883224487304688)
(19, 3.2828211784362793)
(20, 3.326446533203125)
(21, 3.3107962608337402)
(22, 3.37711501121521)
(23, 3.802743673324585)
(24, 3.8378348350524902)
(25, 4.0753912925720215)
(26, 4.123538494110107)
(27, 3.272284507751465)
(28, 2.8650805950164795)
(29, 2.7218852043151855)
(30, 2.4142463207244873)
(31, 2.081291675567627)
(32, 3.257103204727173)
(33, 2.9861817359924316)
(34, 2.767500877380371)
(35, 2.8631765842437744)
(36, 2.793905735015869)
(37, 2.5412588119506836)
(38, 2.437565565109253)
(39, 2.335860013961792)
(40, 2.452831983566284)
(41, 2.4973292350769043)
(42, 2.5277822017669678)
(43, 2.6159191131591797)
(44, 2.581120252609253)
(45, 2.569038152694702)
(46, 2.463578462600708)
(47, 2.492082357406616)
(48, 2.3082497119903564)
(49, 2.2535653114318848)
(50, 2.2331607341766357)
(51, 2.0062527656555176)
(52, 1.9893676042556763)
(53, 2.1211626529693604)
(54, 2.148392677307129)
(55, 2.410344362258911)
(56, 1.8389801979064941)
(57, 2.4110770225524902)
(58, 1.9183589220046997)
(59, 2.0676205158233643)
(60, 2.1850337982177734)
(61, 2.3445074558258057)
(62, 1.0746595859527588)
(63, 1.861149787902832)
(64, 1.541609287261963)
(65, 1.3624505996704102)
(66, 1.7264044284820557)
(67, 1.7814611196517944)
(68, 2.2080771923065186)
(69, 2.787672996520996)
(70, 2.514305830001831)
(71, 2.1176466941833496)
(72, 2.5843493938446045)
(73, 2.63199520111084)
(74, 2.3626177310943604)
(75, 2.0772287845611572)
(76, 2.1602232456207275)
(77, 1.7041916847229004)
(78, 2.7889368534088135)
(79, 2.9629554748535156)
(80, 2.62099289894104)
(81, 2.413357734680176)
(82, 2.258157968521118)
(83, 2.1110050678253174)
(84, 2.4979119300842285)
(85, 2.2642664909362793)
(86, 2.5248234272003174)
(87, 2.510627031326294)
(88, 2.547994375228882)
(89, 3.026803970336914)
(90, 3.1295385360717773)
(91, 2.4916741847991943)
(92, 3.2160327434539795)
(93, 3.1538147926330566)
(94, 3.8139827251434326)
(95, 3.546510934829712)
(96, 3.385956048965454)
(97, 3.2448923587799072)
(98, 3.0440807342529297)
(99, 3.2558224201202393)
(100, 3.188751220703125)
(101, 2.955944538116455)
(102, 2.755589246749878)
(103, 3.736854314804077)
(104, 4.081725597381592)
(105, 3.9731674194335938)
(106, 3.8654301166534424)
(107, 3.8811607360839844)
(108, 3.6713032722473145)
(109, 3.8129734992980957)
(110, 4.147172451019287)
(111, 3.980104446411133)
(112, 3.6912648677825928)
(113, 3.7827699184417725)
(114, 4.98926305770874)
(115, 10.019689559936523)
(116, 7.953170299530029)
(117, 6.81840181350708)
(118, 6.71563196182251)
(119, 6.32946252822876)
(120, 6.420374870300293)
(121, 6.771335601806641)
(122, 6.788459777832031)
(123, 6.805778503417969)
(124, 6.79174280166626)
(125, 6.965189456939697)
(126, 7.702105522155762)
(127, 8.007184982299805)
(128, 7.27396297454834)
(129, 7.341268539428711)
(130, 7.705135822296143)
(131, 8.768766403198242)
(132, 7.304343223571777)
(133, 6.8022637367248535)
(134, 6.649263381958008)
(135, 6.548917293548584)
(136, 13.665461540222168)
(137, 10.239572525024414)
(138, 6.937373161315918)
(139, 6.772017478942871)
(140, 6.932207107543945)
(141, 5.77849006652832)
(142, 5.966350078582764)
(143, 7.025307655334473)
(144, 4.509143829345703)
(145, 4.307125091552734)
(146, 4.325937747955322)
(147, 4.427224636077881)
(148, 3.2614035606384277)
(149, 2.7647457122802734)
(150, 3.8055505752563477)
(151, 2.8146889209747314)
(152, 2.610748291015625)
(153, 2.7973499298095703)
(154, 2.444277048110962)
(155, 2.347437858581543)
(156, 2.6327130794525146)
(157, 2.6717007160186768)
(158, 3.252223014831543)
(159, 2.865025758743286)
(160, 3.370168685913086)
(161, 2.5775139331817627)
(162, 2.4829792976379395)
(163, 3.4727723598480225)
(164, 3.461639881134033)
(165, 3.403855323791504)
(166, 3.6729416847229004)
(167, 3.40334415435791)
(168, 4.118014335632324)
(169, 4.341280460357666)
(170, 3.64516544342041)
(171, 3.8758962154388428)
(172, 3.2758841514587402)
(173, 3.08978009223938)
(174, 3.1280691623687744)
(175, 3.0988118648529053)
(176, 3.0013744831085205)
(177, 2.847583532333374)
(178, 2.7288284301757812)
(179, 2.1070926189422607)
(180, 1.9359793663024902)
(181, 1.4179885387420654)
(182, 1.3984953165054321)
(183, 1.754085659980774)
(184, 2.1874005794525146)
(185, 4.148775100708008)
(186, 3.263681411743164)
(187, 2.576789140701294)
(188, 2.3193609714508057)
(189, 2.3541762828826904)
(190, 2.247575521469116)
(191, 2.2071633338928223)
(192, 2.8883049488067627)
}

\newcommand{\drawAdaptiveFrameWise}{%
(0, 3.359450101852417)
(1, 3.4269168376922607)
(2, 3.437886953353882)
(3, 3.0076379776000977)
(4, 3.3969178199768066)
(5, 3.11140775680542)
(6, 3.1027820110321045)
(7, 3.628429651260376)
(8, 3.188166856765747)
(9, 3.117325782775879)
(10, 2.9738290309906006)
(11, 3.085413932800293)
(12, 3.385896921157837)
(13, 3.695819854736328)
(14, 3.7226040363311768)
(15, 3.3165807723999023)
(16, 3.1772968769073486)
(17, 3.3839914798736572)
(18, 3.377108573913574)
(19, 3.4179575443267822)
(20, 3.4365272521972656)
(21, 3.3025362491607666)
(22, 3.079322338104248)
(23, 3.4273877143859863)
(24, 3.4967610836029053)
(25, 4.037619590759277)
(26, 3.806363344192505)
(27, 3.761291265487671)
(28, 3.8606369495391846)
(29, 4.342137813568115)
(30, 3.1518473625183105)
(31, 2.9852981567382812)
(32, 4.363644123077393)
(33, 3.0081632137298584)
(34, 2.697706937789917)
(35, 2.60477876663208)
(36, 2.4250247478485107)
(37, 2.313491106033325)
(38, 2.764230489730835)
(39, 2.7525947093963623)
(40, 2.808987617492676)
(41, 2.6718685626983643)
(42, 2.6528077125549316)
(43, 2.680190324783325)
(44, 2.543663740158081)
(45, 2.5487310886383057)
(46, 2.5178582668304443)
(47, 2.620347499847412)
(48, 2.335465908050537)
(49, 2.2850873470306396)
(50, 2.382938861846924)
(51, 2.1762564182281494)
(52, 2.1251819133758545)
(53, 2.0013175010681152)
(54, 1.909181833267212)
(55, 1.7802809476852417)
(56, 3.2805886268615723)
(57, 1.412118673324585)
(58, 0.6775956749916077)
(59, 1.2987333536148071)
(60, 1.1619185209274292)
(61, 0.8304275870323181)
(62, 2.5150139331817627)
(63, 1.7921159267425537)
(64, 1.2488356828689575)
(65, 1.7012430429458618)
(66, 1.6833539009094238)
(67, 1.4282135963439941)
(68, 1.4299957752227783)
(69, 0.8536927103996277)
(70, 4.046439170837402)
(71, 3.7517542839050293)
(72, 2.2753777503967285)
(73, 2.557086229324341)
(74, 3.309431791305542)
(75, 4.662275791168213)
(76, 5.340081691741943)
(77, 3.1087892055511475)
(78, 4.545870304107666)
(79, 2.1645491123199463)
(80, 2.199716091156006)
(81, 2.6179122924804688)
(82, 2.4115495681762695)
(83, 2.505880355834961)
(84, 3.025409698486328)
(85, 2.6346280574798584)
(86, 3.4433581829071045)
(87, 2.891714572906494)
(88, 3.0684919357299805)
(89, 3.5193774700164795)
(90, 3.161747932434082)
(91, 3.0856587886810303)
(92, 4.586146831512451)
(93, 3.2557871341705322)
(94, 3.4959499835968018)
(95, 3.420588731765747)
(96, 3.631337881088257)
(97, 3.556871175765991)
(98, 3.56494140625)
(99, 4.245471000671387)
(100, 4.109501361846924)
(101, 3.856147050857544)
(102, 4.463545322418213)
(103, 5.61824369430542)
(104, 5.037607669830322)
(105, 3.8230628967285156)
(106, 3.8580126762390137)
(107, 4.307868003845215)
(108, 4.443691730499268)
(109, 5.25106143951416)
(110, 5.375445365905762)
(111, 4.582160472869873)
(112, 4.725798606872559)
(113, 4.95527458190918)
(114, 7.581338882446289)
(115, 17.28974723815918)
(116, 2.879368543624878)
(117, 1.8587003946304321)
(118, 2.5490918159484863)
(119, 3.2824411392211914)
(120, 2.80880069732666)
(121, 3.753148078918457)
(122, 2.873410701751709)
(123, 3.302525758743286)
(124, 3.0873823165893555)
(125, 3.313338041305542)
(126, 7.719971656799316)
(127, 2.9259836673736572)
(128, 2.091752052307129)
(129, 2.8502047061920166)
(130, 3.3423402309417725)
(131, 15.382074356079102)
(132, 3.071636438369751)
(133, 3.0186874866485596)
(134, 3.4395461082458496)
(135, 5.156672954559326)
(136, 19.441062927246094)
(137, 5.185881614685059)
(138, 6.454106330871582)
(139, 3.396317481994629)
(140, 2.52252459526062)
(141, 3.414276123046875)
(142, 1.9009158611297607)
(143, 1.8081555366516113)
(144, 7.724588394165039)
(145, 3.805584669113159)
(146, 1.5994107723236084)
(147, 1.3144103288650513)
(148, 1.5766032934188843)
(149, 1.775449514389038)
(150, 1.8882544040679932)
(151, 2.0433082580566406)
(152, 21.391611099243164)
(153, 3.6497814655303955)
(154, 2.829777717590332)
(155, 2.968831777572632)
(156, 3.0994813442230225)
(157, 3.1783182621002197)
(158, 3.1691648960113525)
(159, 3.2887911796569824)
(160, 4.032301425933838)
(161, 3.316749334335327)
(162, 4.329782009124756)
(163, 4.264927387237549)
(164, 3.6303822994232178)
(165, 3.7536418437957764)
(166, 5.574972629547119)
(167, 4.869246006011963)
(168, 5.960879802703857)
(169, 6.599457263946533)
(170, 9.809332847595215)
(171, 3.850440740585327)
(172, 3.2849416732788086)
(173, 3.1487603187561035)
(174, 3.1611826419830322)
(175, 3.141291856765747)
(176, 3.0831546783447266)
(177, 2.940117120742798)
(178, 2.846896171569824)
(179, 2.9197580814361572)
(180, 6.144878387451172)
(181, 4.436007976531982)
(182, 3.756174325942993)
(183, 3.7117574214935303)
(184, 4.339858531951904)
(185, 6.772038459777832)
(186, 2.1343302726745605)
(187, 1.7536050081253052)
(188, 2.318730354309082)
(189, 2.8491687774658203)
(190, 2.961761236190796)
(191, 3.8558592796325684)
(192, 5.032707691192627)
}

\usepgfplotslibrary{fillbetween}
\pgfplotsset{compat=1.3}
\pgfplotsset{xtick style={draw=none}}
\pgfplotsset{ytick style={draw=none}}

\pgfplotsset{major grid style={gray!40}}
\pgfdeclarelayer{background}
\pgfdeclarelayer{foreground}
\pgfsetlayers{background,main,foreground}
\definecolor{C0}{rgb}{0.121569, 0.466667, 0.705882}
\definecolor{C1}{rgb}{1.000000, 0.498039, 0.054902}
\definecolor{C2}{rgb}{0.172549, 0.627451, 0.172549}
\definecolor{C3}{rgb}{0.839216, 0.152941, 0.156863}
\definecolor{C4}{rgb}{0.580392, 0.403922, 0.741176}
\definecolor{C5}{rgb}{0.549020, 0.337255, 0.294118}
\definecolor{C6}{rgb}{0.890196, 0.466667, 0.760784}
\definecolor{C7}{rgb}{0.498039, 0.498039, 0.498039}
\definecolor{C8}{rgb}{0.737255, 0.741176, 0.133333}
\definecolor{C9}{rgb}{0.090196, 0.745098, 0.811765}
\definecolor{C10}{rgb}{0.337255, 0.341176, 0.333333}
\definecolor{C11}{rgb}{0.590196, 0.745098, 0.811765}

\newcommand{\cs}{}
\newcommand{\hh}{0mm}
\newcommand{\hhh}{0mm}
\newcommand{\hhhh}{0mm}
\newcommand{\vvv}{0mm}
\newcommand{\vvvv}{0mm}

\newcommand{\vlabel}[3]{\makebox[0mm][l]{\rotatebox{90}{\makebox[#2][c]{#3}}}\hspace{#1}}
\newcommand{\hrlabel}[1]{\hfill\makebox[0mm]{#1}\hfill}
\newcommand{\vrlabel}[1]{\vfill\makebox[0mm]{#1}\vfill}

\begin{abstract}
Shape-from-Template (SfT) refers to the class of methods that reconstruct the 3D shape of a deforming object from images/videos using a 3D template. Traditional SfT methods require point correspondences between images and the texture of the 3D template in order to reconstruct 3D shapes from images/videos in real time. Their performance severely degrades when encountered with severe occlusions in the images because of the unavailability of correspondences. In contrast, modern SfT methods use a correspondence-free approach by incorporating deep neural networks to reconstruct 3D objects, thus requiring huge amounts of data for supervision. Recent advances use a fully unsupervised or self-supervised approach by combining differentiable physics and graphics to deform 3D template to match input images. In this paper, we propose an unsupervised SfT which uses only image observations: color features, gradients and silhouettes along with a mesh inextensibility constraint to reconstruct at a $400\times$ faster pace than (best-performing) unsupervised SfT. Moreover, when it comes to generating finer details and severe occlusions, our method outperforms the existing methodologies by a large margin. Code is available at \href{https://github.com/dvttran/nsft}{https://github.com/dvttran/nsft}.
\end{abstract}    
\section{Introduction}
\label{sec:intro}

\begin{figure}
    \centering
    \includegraphics[width=1.0\linewidth]{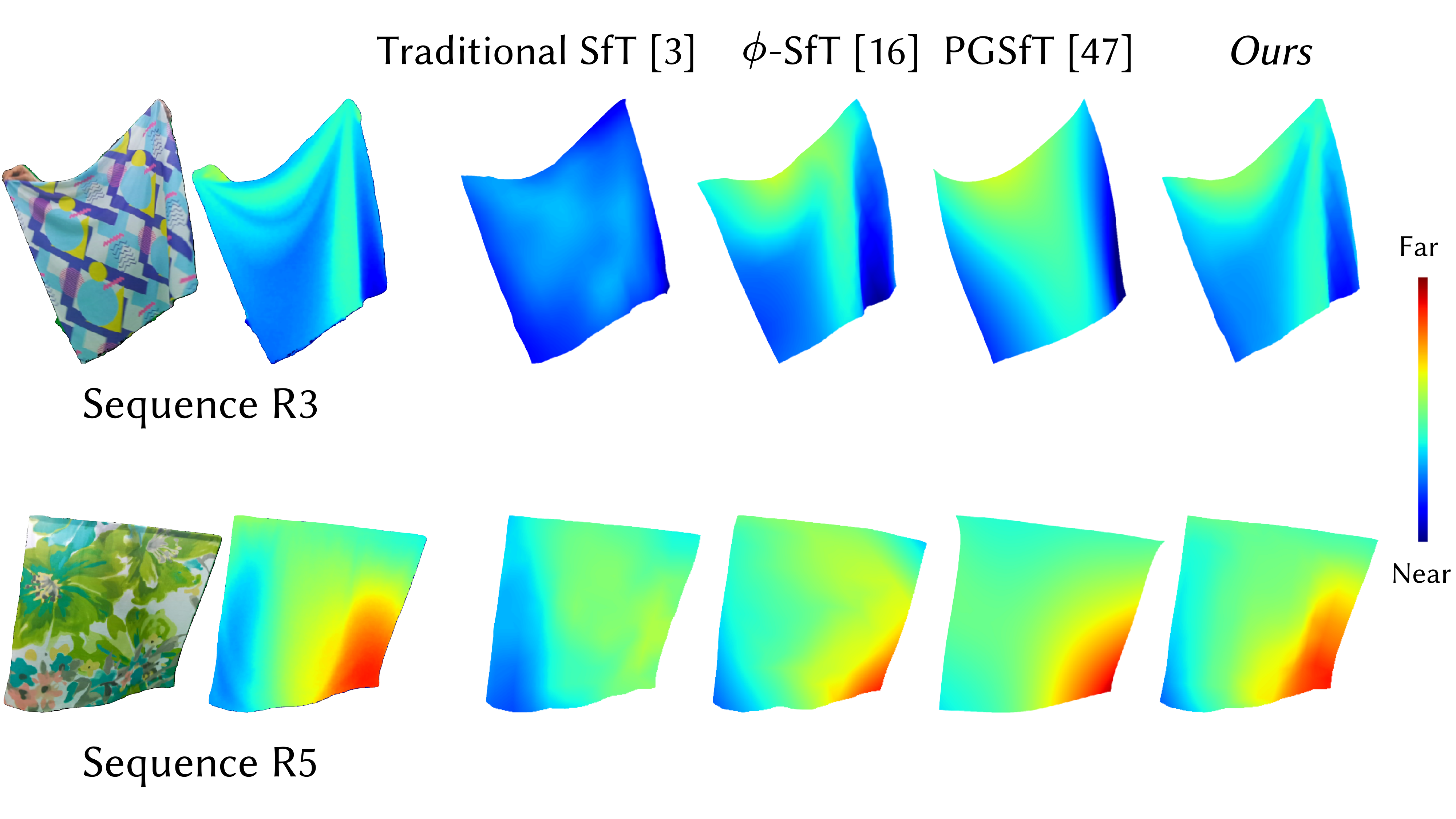}
    \caption{Image-guided SfT vs SoTA. 
    Our method recovers finer details, resolves self-occlusions and handles strong perspectivities better than existing methods. We evaluated traditional SfT \cite{Bartoli2015} with correspondences obtained from Cotracker v3~\cite{karaev2024cotracker3}.}
    \label{fig:teaser}
\end{figure}

Recovering 3D shapes of objects from monocular RGB images, obtained through a video sequence or wide-baseline viewpoints, is a key goal in 3D computer vision. Given a 3D template of an object, comprising a known geometric configuration of the object  along with known texture details,~\cite{Bartoli2015,Salzmann2008} established that a unique 3D shape can be recovered from a single image of an object in a configuration different from the 3D template. It showed that combining projective geometry of the calibrated images with isometric (or geodesic-preserving) constraints on the 3D template deformation, the underlying 3D structure of any point in the image can be uniquely identified merely from the point-wise registration between the input image and the texture map associated with the 3D template. Template-based 3D reconstruction from images, a.k.a SfT, is therefore proven to be a well-posed problem with reliable solutions~\cite{Chhatkuli2016} that can perform real-time 3D reconstructions~\cite{Collins2015,Ngo2016,Shetab2024,Tan2024}.
~\cite{Ngo2015} extends the traditional SfT to adapt to handle occlusions by interpolating the non-corresponding points on images from the reconstructed 3D points of the relevant neighbors. 
However, all these methods fail when the deforming object causes large (self-)occlusions, sharp motion, and high perspectivities in the images, such as in the case of the deforming cloth shown in~\Cref{fig:teaser}. This is because a unique solution of 3D shape is possible only if the point correspondences between the template texture map and the input images are known. Such occlusion artifacts appear quite often while capturing deforming objects; thus the inefficacy of traditional SfT to process them poses a severe limitation to their applicability. Sharp motions and strong perspectivities also significantly degrade the reliability of point correspondences which causes these methods to fail. To mitigate this issue, several modern SfT methods~\cite{Fuentes2021,Fuentes2022,Pumarola2018,Golyanik2018,Shimada2019,Sanchez2025} use deep neural networks (DNN) that can jointly perform template-image registration and 3D reconstruction. These methods may require a large amount of data for supervision and/or several restrictions on the scene and camera. However, they fail to recover finer details and severe occlusions.  

Alternatively, $\phi$-SfT~\cite{Kairanda2022} proposed to take advantage of the physics-based simulation of thin-shell objects~\cite{Liang2019,Narain2012} to deform a 3D template and render using differentiable graphics to match input images in a complete unsupervised fashion. Although successful in handling fine details and severe occlusions to some extent, it incurs a huge computational cost, which makes it practically limited. Recently, PGSfT~\cite{Stotko2024} used self-supervised learning of physics-based thin-shell simulations~\cite{Santesteban22,Chen2024} to learn a neural cloth model which allowed a $400\times$ speedup over~\cite{Kairanda2022}, although with a degraded performance in handling severe occlusions and generating fine details.  

In this paper, like~\cite{Kairanda2022}, we present a correspondence-free, unsupervised SfT technique to reconstruct 3D shapes. However, instead of using physics-based simulation of thin shell objects, we rely on simple image-related measures only to guide the 3D template mesh deformations. Consequently, we achieve a $400\times$ computational speed up with a significantly improved performance than~\cite{Kairanda2022}, especially while recovering finer details and challenging motion. Moreover, we initialize the simulation of a given video frame with its previous one, allowing a simpler and faster prediction of the current shape. Furthermore, we enforce an inextensibility measure of the template mesh rather than a strict isometric deformation~\cite{Bartoli2015} to allow the template to change its geometry according to the deformations observed in the images. Such a non-strict measure allows us to deal with isometric objects such as a paper and elastic objects such as clothes equally well. Our experiments show that the proposed method outperforms traditional and modern methodologies by large margins, especially while reconstructing finer details and severe occlusions.

\section{Related Works}
\label{sec:sota}

\noindent \textbf{Traditional SfT methods.} Given the registration between images and the texture map of the 3D template, these methods reconstruct surfaces as seen in input images assuming certain constraints on the deformations.~\cite{Salzmann2008,Salzmann2009,Perriollat2011,Brunet2014} assume that the template is represented with a mesh and enforce inextensibility constraints on its deformations to reconstruct the surfaces.~\cite{Bartoli2015,Chhatkuli2016} assume a continuous representation of the template and enforce strict isometric constraints on its deformations. The reconstructions are obtained as a solution to a Partial Differential Equation. These methodologies have been extended to other deformation models such as conformality~\cite{Bartoli2015}, equiareality~\cite{Casillas2019,Parashar2020}, ARAP~\cite{Parashar2015}, smoothness~\cite{Ozgur2017} and elasticity~\cite{Malti2013,Malti2015,Malti2017,Haouchine2017,Agudo2015} in order to incorporate more realistic deformation scenarios. These methods are mostly computationally efficient and perform well in generic scenarios, but they are prone to high errors when encountered with strong deformations, severe occlusions, and lighting changes in the images. Under these conditions, the point correspondences obtained from the registration are highly inaccurate which contributes strongly to the degradation in their performance.~\cite{Yin2016,Ngo2016} attempted to jointly estimate the shape and registration in order to overcome these issues; however, they could not achieve any significant performance improvement.

\noindent \textbf{DNN-based SfT methods.} 
~\cite{Pumarola2018,Golyanik2018} consider the template to be a flat, regular mesh and estimate the displacement of 3D vertices to comply with observed images using an encoder-decoder network. These methods learn object-specific models that rely on both registration and reconstructions as ground truth for the supervision. Such data is difficult to obtain, and the learned models are limited to the images seen in training scenarios for which registration could be computed. Therefore, these methods do not generalize well and do not handle complex deformation scenarios.~\cite{Fuentes2022} proposed an object-specific model that can be learned without using registration data as labels and is not limited to flat surfaces. It trains an encoder-decoder network from a large amount of deformed shapes (typically 100K samples) of the given object.  It can reconstruct volumetric objects as well using an additional RGBD sensor as an input to monitor the non-visible surface. It can generalize better to unseen object configurations than~\cite{Pumarola2018,Golyanik2018}, but cannot handle severe occlusions and lighting changes.~\cite{Sanchez2025} simplifies~\cite{Fuentes2022} by restricting the training to learn registration. Instead of relying on ground truth optical flow for training as in~\cite{Pumarola2018,Golyanik2018}, it uses an off-the-shelf optical flow method to calculate training registration and imposes external losses to control the quality of tracks. 
~\cite{Shimada2019,Fuentes2021} extend~\cite{Golyanik2018} and~\cite{Fuentes2022} respectively by training networks with fixed object geometry but variable textures; thus improving the models from strictly object-specific to only geometry-specific. However, this does not significantly improves performance.
All these methods cannot handle strong deformations, sharp motions, severe occlusions and fail to generate finer details.
    
Inspired by physics-based simulation~\cite{Liang2019,Narain2012,Qiao2020,Li2022} and recent advances in physics recovery from videos~\cite{Jacques2020,Guo2021,Li2021,Weiss2020,Rempe2020},~\cite{Kairanda2022} proposed to solve SfT using differentiable physics and graphics to recover template deformation/motion according to input images in an unsupervised fashion. This approach does not require/compute registration, and recovers a decent 3D shape even in the cases of strong deformations and severe occlusions. A huge downside is that it is prohibitively slow: it can take upto 30 hours to process 50-60 images. Measures have been proposed to speed-up the computation; however they are based upon reducing the quality of 3D template mesh which significantly affects the quality of reconstruction. Using the strategy proposed in~\cite{Baraff1998} to transform physics-based simulation into an optimisation problem,~\cite{Santesteban22,Chen2024} perform a self-supervised learning of clothed garments under various motions.~\cite{Stotko2024} adapted this methodology to solve SfT using self-supervised learning of template deformations. It learns a mesh-specific and a motion-specific model with a $400\times$ speedup over~\cite{Kairanda2022}; however at a degraded performance.

In this paper, we simplify the unsupervised physics-based simulation in~\cite{Kairanda2022} to achieve a $400\times$ speedup  while outperforming it by large margins, even in complex deformations scenario.  Instead of estimating the physical parameters related to stress, strain and bending energies, we pose the template deformation as an offset prediction of template mesh's vertices. This is a widely used approach in computer graphics~\cite{Chen2024,Santesteban22}. Instead of directly manipulating the physical forces to achieve equilibria in the above-mentioned energies, we rely solely on image observations to indirectly find the equilibrium matching the input images. In addition, we force mesh inextensibility on the template as a regularizer. It allows us to deal with both isometric and elastic surfaces, allowing mesh extensions pertaining to input images.

\noindent \textbf{Summary of contributions.} 

1) We propose a simple, image-guided framework to simulate deformations using only vision cues (color, edges, gradients) to recover the motion as opposed to explicit estimation of physical properties.

2) We model deformations with mesh offset prediction using a deformation network. It is a compact, low-dimensional module that captures the essential deformations while naturally leaving out redundant learning variables associated with explicit physics-based learning.

3) We use an adaptive data loss structure to compute vision losses which allows us to effectively handle shading variations arising due to deformations. It is crucial for reconstructing finer details. 

4) We propose a simplified optimization strategy that reduces computational complexity from $O(T^2N)$ in~\cite{Kairanda2022,Stotko2024} to $O(TN)$, where $T$ is the number of frames in the length of the video sequence and $N$ is average number of iterations needed to reconstruct a single frame. Moreover, our frame-wise optimization implicitly enforces temporal coherence rather than optimizing temporal losses additionally.

5) Our method outperforms SoTA by large margins in terms of speed and accuracy. It handles complex deformation scenarios extremely well. We show the strengths of our method, optimization strategy and data loss structure with extensive evaluation.

\section{Image-guided SfT}
\label{sec:method}
\begin{figure*}
    \centering
    \includegraphics[width=0.9\linewidth]{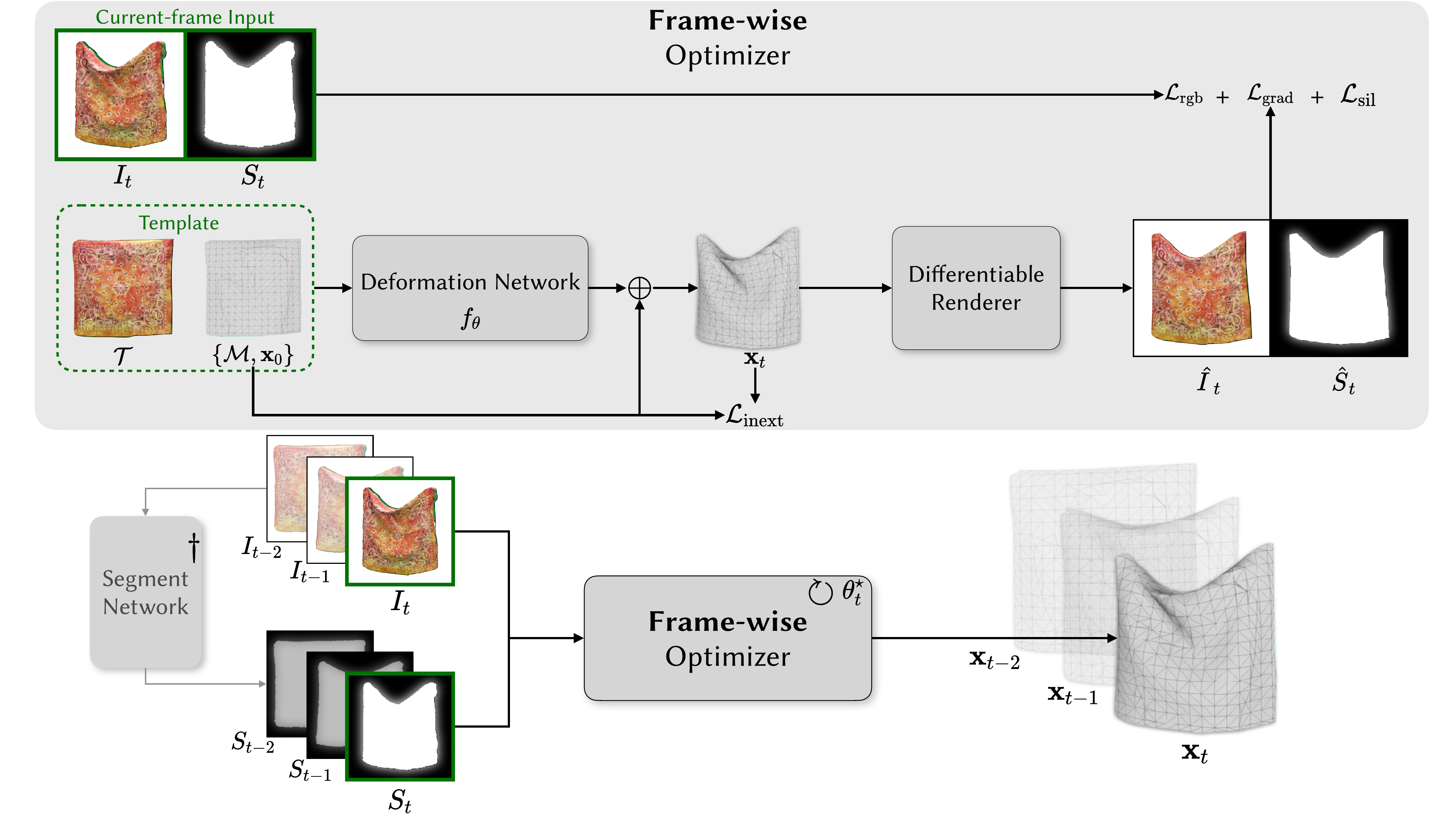}
    \caption{\textbf{Overview of our frame-wise image-guided SfT pipeline.} The template is provided with a texture map $\mathcal{T}$, the mesh connectivity $\mathcal{M}$ and the initial shape $\mathbf{x}_0$. We learn the reconstruction on a video sequence frame-by-frame. At current frame $t$, the deformation network $f_\theta$ predicts a displacement from $\mathbf{x}_0$ to produce the deformed shape $\mathbf{x}_t$. A differentiable renderer then projects this mesh to produce a rendered RGB image $\hat{I}_t$ and silhouette $\hat{S}_t$. We compute pixel‐wise vision losses and a mesh inextensibility regularization, and optimize for the best parameters $\theta_t^\star$ via backpropagation. These parameters are passed forward to initialize the next frame’s optimization (the symbol $\circlearrowright$). $^\dagger$If ground-truth is not available, an off‐the‐shelf segmentation network can be employed to generate the reference silhouette masks.} 
    \label{fig:pipeline}
\end{figure*}

Figure~\ref{fig:pipeline} shows our pipeline.
We consider template as a textured triangle mesh $\mathcal{M}=(\mathcal{V},\mathcal{E},\mathcal{F},\mathcal{T})$ with topological connectivity given by the edges $\mathrm{e}\in\mathcal{E}$, the triangular faces $\mathrm{f}\in\mathcal{F}$ and texture map $\mathcal{T}$. The geometry of the mesh is the set of vertex coordinates $\mathbf{x}=\{x_\mathrm{v}\in\mathbb{R}^3\mid\mathrm{v}\in\mathcal{V}\}\in\mathbb{R}^{|\mathcal{V}|\times 3}$. We consider that the 3D shape at the initial frame is a rigid transformation of the template.  We denote $\mathbf{x}_0$ for the template and $\mathbf{x}_t$, for $t=1,\dots,T$ the shapes to be reconstructed.
In addition, we assume that the video input is accompanied by the camera calibration, the intrinsic matrix $\mathbf{K}\in\mathbb{R}^{3\times 3}$. Since we are reconstructing as per camera frame, we do not require camera extrinsics to be known. 

\subsection{Deformation Model}\label{sec:deformation}
Surface moves according to a deformation field $\varphi_t:\mathbb{R}^3\to\mathbb{R}^3$. To reconstruct a surface shape, we need to learn such a vertex-wise mapping and denote $\mathbf{x}_t:=\varphi(\mathbf{x}_0;\mathbf{z}_t)$ for some latent variable $\mathbf{z}_t$. Unlike physics-based methods that explicitly optimize the physics parameters, e.g. scene or material conditions, etc., we parametrize the surface deformation via a \emph{deformation network} $f_{\theta}(\mathbf{x}_0,t)$ such that
\begin{align}
    \mathbf{x}_{t}=\mathbf{x}_{0}+f_{\theta}(\mathbf{x}_0,t).
\end{align}
Such a neural network inherently provides a continuous mapping from the domain of vertex coordinates (and time) to their corresponding displacements. Moreover, it allows us to transfer the learning in $\mathbf{z}_t$ to learning the network parameters, which enables an adjustable balance between expressivity and computational efficiency by tuning the network architecture. We refer to Appendix \ref{appendix:deformation-network} for discussion related to the advantages of our choice of deformation modeling with respect to the well established ones, such as vertex offset prediction, physics-based modeling, etc. 

%
%
%

\subsection{Differentiable Renderer}

Once the camera calibration is known, we can utilize off-the-shelf differentiable renderers \cite{Ravi2020, Laine2020} to optimize the shapes by aligning the vision cues within video sequence. We can see such renderers as differentiable functions that map the shape $\mathbf{x}_t$ to an RGB image $\hat{I}_t$ and a silhouette $\hat{S}_t$. We note that the renderers require the mesh to be textured, i.e. a texture image $\mathcal{T}$ and its corresponding UV map in the normalized coordinate space $[0,1]^2$ are available. 

\subsection{Data Loss Structure}\label{sec:adaptive-data-loss}
Before diving into the explicit loss functions, we explain the data loss structure common to all vision losses. We express the \textit{adaptive} data loss between a pair of prediction and ground-truth data as
\begin{align}
    \ell(\hat{y},y)=w(|\hat{y}-y|)\odot\ell_p(\hat{y},y),\label{eq:data-loss-1}
\end{align}
where $\odot$ denotes the element-wise product, $\ell_p$ is the $\ell$-$p$ norm and the weighting factor
\begin{align}
    w(d)= \alpha \exp\left(\dfrac{d}{\sigma}\right),\label{eq:data-loss-2}
\end{align}
controlled by the hyper-parameters $\alpha$ and $\sigma$. This scheme amplifies element-wise errors exponentially, guiding the optimizer to focus on significant discrepancies. 

Such an adaptive loss is particularly useful for our vision losses. In practice, many renderers either neglect or approximate lighting effects, while real objects exhibit dynamic variations in brightness across video frames. Thus, the rendered surface can visually differ from the ground truth. However, the color mismatch typically remains more significant than any lighting discrepancies, allowing our adaptive loss to effectively handle these shading variations without sacrificing performance on regions with large errors. 

\subsection{Loss Functions}
Like \cite{Kairanda2022}, we optimize the shapes using two main losses on RGB images and silhouettes (i.e., masks of the object in the frames). For in-the-wild videos, where the silhouettes might not be available, we propose to use Segment Anything Model 2 (SAM 2) \cite{ravi2024sam2} that can generate the object masks in images and videos from user prompts as point click or bounding box, etc. %


\noindent \textbf{RGB Loss.}
It measures the pixel-wise difference between the rendered frame $\hat{{I}}_t$ and the video frame $\tilde{I}_t$ that is masked to isolate the object of interest, has the form
\begin{align}
    \mathcal{L}_\text{rgb}(\mathbf{x}_t)=\ell(\hat{{I}}_t,I_t).
\end{align}
\noindent \textbf{Silhouette Loss.}
The object position within the frame is further constrained by silhoutte loss. Using data loss $\ell$ as above, we define the silhouette loss as
\begin{align}
    \mathcal{L}_\text{sil}(\mathbf{x}_t)=\ell(\hat{{S}}_t,S_t),
\end{align}
where $\hat{{S}}_t$ is the rendered silhouette and $S_t$ denotes the ground-truth mask.
A Gaussian blur filter is also applied for smooth gradients on the boundaries, as in \cite{Kairanda2022}. 

\noindent \textbf{Image Gradient Loss.}
To exploit more information from the video, we propose to use the image gradient loss
\begin{align}
    \mathcal{L}_\text{grad}(\mathbf{x}_t)=\ell(k(\hat{I}_t),k(I_t)),
\end{align}
where $k(\cdot)$ denotes an image gradient operator (e.g., Sobel). By extracting first-order information from each image, this operator highlights edges and localized intensity variations, which can be crucial for guiding surface reconstructions on well-textured objects.

\subsection{Mesh Inextensibility Regularization}\label{sec:isometry}
To keep the optimization more stable, we use a mesh inextensibility introduced by \cite{Chen2024}, of the form
\begin{align}
    \mathcal{L}_\text{inext}(\mathbf{x}_t)=w_\text{inext}\sum_{\mathrm{v}\in\mathcal{V}}\lvert\text{det}(\hat{C}_{t,\textrm{v}}-\Lambda_{0,\textrm{v}})\rvert,\label{eq:inextensibility-loss}
\end{align}
where $\hat{C}_{t,\textrm{v}}$ represents the predicted covariance matrix at vertex $x_{t,\mathrm{v}}$ about its neighborhood $\mathcal{N}_{t,\mathrm{v}}$ with mean $\bar{x}_{t,\mathrm{v}}$ given by
\begin{align}
    \hat{C}_{t,\textrm{v}}=\dfrac{1}{|\mathcal{N}_{t,\mathrm{v}}|}\sum_{\mathrm{n}\in\mathcal{N}_{t,\mathrm{v}}}(x_{t,\mathrm{n}}-\bar{x}_{t,\mathrm{v}})(x_{t,\mathrm{n}}-\bar{x}_{t,\mathrm{v}})^\top,\label{eq:covariance-matrix}
\end{align}
and the matrix $\Lambda_{0,\textrm{v}}$ is derived from the eigenvalues of the template covariance matrix $C_{0,\textrm{v}}$ as
\begin{align}
    \Lambda_{0,\mathrm{v}}=\begin{bmatrix}
    \lambda_{0,\mathrm{v}}^{1} & 0 & 0 \\
    0 & \lambda_{0,\mathrm{v}}^{2} & 0 \\
    0 & 0 & \lambda_{0,\mathrm{v}}^{3} \\
    \end{bmatrix}.\label{eq:template-covariance-matrix}
\end{align}
The weighting factor $w_\text{inext}$ is computed adaptively with respect to the mesh scale. 

\subsection{Frame-wise Optimization}\label{sec:frame-wise-optimization}
As discussed in Section \ref{sec:deformation}, our goal is to learn the optimal parameters of the deformation network to reconstruct the shape $\mathbf{x}_t$, using the loss functions and regularization above. 
Our main idea is to reconstruct the object shapes for each frame independently. In particular, for each frame $t$, we seek the parameters $\theta_t^\star$ such that the reconstructed surface ${\mathbf{x}}_t=\mathbf{x}_0+f_{\theta_t^\star}(\mathbf{x}_0,t)$ minimizes the losses aforementioned. Once attained, the parameters $\theta_t^\star$ is passed as the initialization for the next frame, $\theta_{t+1}^{(0)}=\theta_t^\star$. This choice benefits from temporal continuity in video data, where consecutive frames tend to have similar displacements, thus requiring relatively small adjustments to the network.
This frame-wise scheme is the key to our significant speedup, especially for long video sequences. More details on complexity are described in Appendix \ref{appendix:complexity-analysis}.




\section{Experiments}
\label{sec:evaluation}

We implemented our framework in \emph{PyTorch} \cite{paszke2019pytorch}, leveraging automatic differentiation to use gradient-based optimization to learn the deformation network, and the \emph{libigl} \cite{libigl} library for geometry processing on mesh. We used the \emph{nvdiffrast} \cite{Laine2020} to  render as it is faster than \emph{PyTorch3D} \cite{Ravi2020}. 

We specify here the default setting for most of our experiments. Our base network is a simple multi-layer perceptron (MLP) with ReLU activations and 8 hidden layers of width 256 for all video sequences. The frame-wise optimization strategy is employed with AdamW optimizer \cite{loshchilov2017decoupled}, learning rate of $10^{-4}$ and weight decay of $10^{-2}$. The learning process warms up with 500 iterations and then runs 200 iterations for each frame. The data loss in Equations \eqref{eq:data-loss-1} and \eqref{eq:data-loss-2} is used for $\mathcal{L}_\text{rgb}$, $\mathcal{L}_\text{sil}$ and $\mathcal{L}_\text{grad}$ with $\alpha=10$ and $\sigma=1$. The Sobel operator is used on both first- and second-order to compute $\mathcal{L}_\text{grad}$ using the \emph{Kornia} library \cite{riba2020kornia}. The experiments are evaluated on a single NVIDIA V100 GPU.

We evaluated our method on Kinect-paper dataset~\cite{varol2012constrained} with 193 images and depth maps recorded using Kinect v1. We used SAM2 to generate image masks. We also experimented with $\phi$-SfT dataset \cite{Kairanda2022}. It has 4 synthetic sequences  of cloth deformations simulated using ArcSim~\cite{Liang2019} and 9 real sequences cloth deformation with various shapes and textures recorded from Azure Kinect v2 camera. The texture and template are provided as 3D surface corresponding to first frame in the sequence. 
Then, we provide an ablation study on various choices of attributes and architectures used in our method.

\subsection{Evaluation on Kinect Paper Dataset}
\begin{figure}
    \centering
    \includegraphics[width=1.\columnwidth]{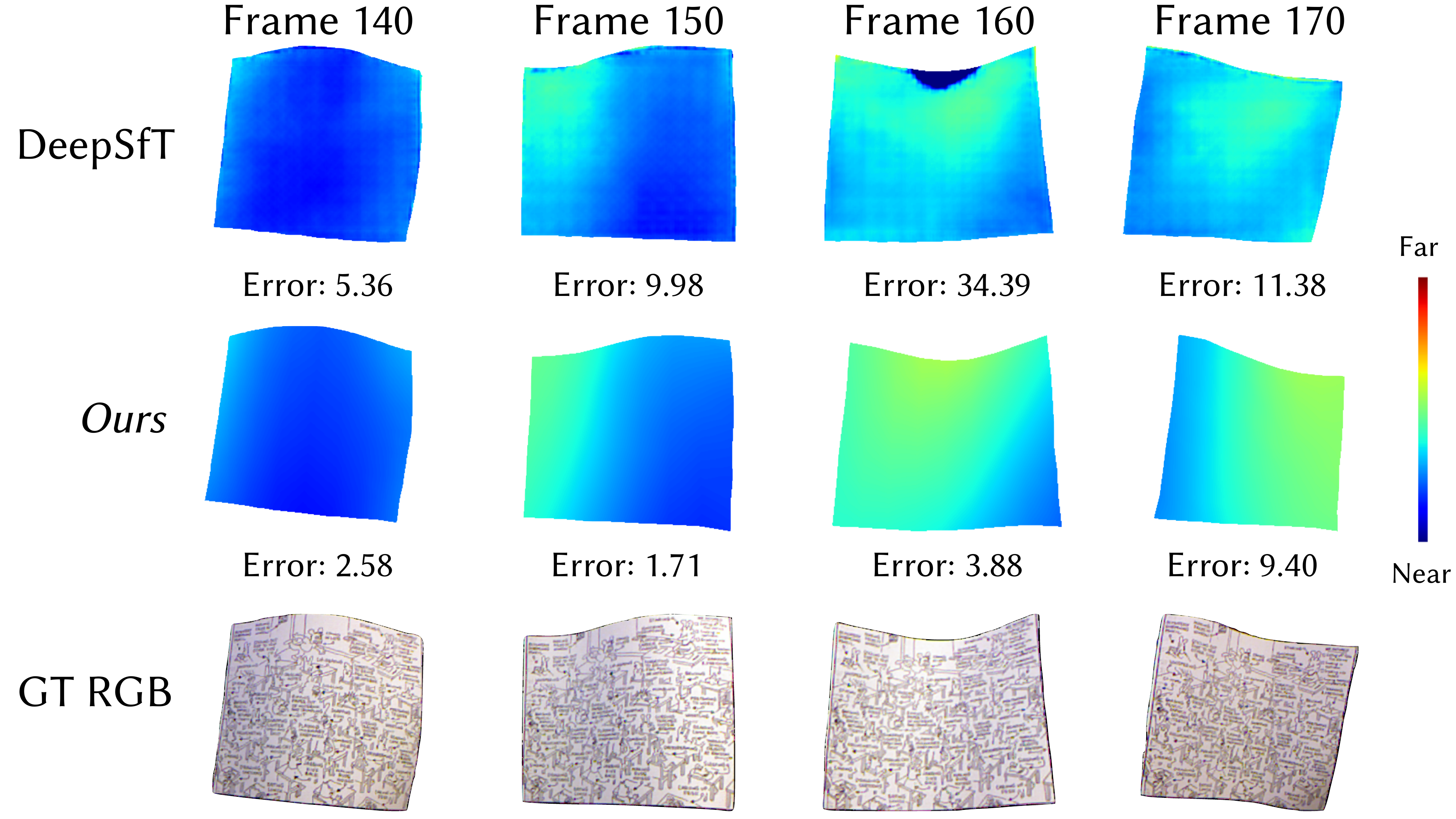}
    \caption{Visual comparison of DeepSfT vs \emph{Ours} on Kinect Paper dataset. Depth map error at each frame is reported in mm.}
    \label{fig:kinect_paper}
\end{figure}

\Cref{tab:kinect_paper} compares quantitative results for traditional SfT~\cite{Bartoli2015}, supervised SfT: DeepSfT~\cite{Fuentes2022} and TD-SfT~\cite{Fuentes2021}, and unsupervised $\phi$-SfT~\cite{Kairanda2022}, with our method outperforming DeepSfT and traditional SfT.  We did not compute results for $\phi$-SfT~\cite{Kairanda2022} due to its high computation time. 
PGSFT~\cite{Stotko2024} is designed to work with square meshes only. We could not evaluate Kinect-paper sequence on it. Selected reconstructed frames are shown in~\Cref{fig:kinect_paper}. The code for TD-SfT~\cite{Fuentes2021} is not publicly available, so we cannot show visual results. While DeepSfT~\cite{Fuentes2022} and TD-SfT~\cite{Fuentes2021} have almost real-time performance, \emph{Ours} takes almost 2 seconds to process each frame. The traditional SfT~\cite{Bartoli2015} takes almost 10 seconds per frame. The correspondences were computed using CoTracker v3 \cite{karaev2024cotracker3}.

 \begin{table}
   \centering
   \caption{Depth map RMSE comparison on Kinect Paper dataset.}
   \resizebox{1\linewidth}{!}{
   \begin{tabular}{rcccc}
      \toprule
     & DeepSfT \cite{Fuentes2022} & TD-SfT \cite{Fuentes2021} & SfT~\cite{Bartoli2015}   & \emph{Ours} \\
     \midrule
     RMSE (mm)  & 6.97 & 3.37 & 6.17  & 4.01 \\
      \bottomrule
   \end{tabular}
   }
   \vspace{3pt}
   \label{tab:kinect_paper}
 \end{table}

\subsection{Evaluation on Synthetic $\phi$-SfT Dataset}
We evaluated our method on 4 sequences of this dataset and compared with $\phi$-SfT and PGSFT. The results are shown in~\Cref{tab:kinect_paper}. The errors are computed as relative 3D error between the ground truth and the reconstructed mesh. We could not run  the supervised method, DeepSfT on this sequence as it requires to be trained separately on each sequence and the training requires almost 100K samples of simulated data. These sequences each have 50 frames only, which are insufficient to train DeepSfT. In the papers, DeepSfT and TD-SfT report limitations with large occlusions and have not shown any experiments with complex geometries. A recent, weakly-supervised SfT~\cite{Sanchez2025} also requires more than 50 frames to train. Thus, we do not make further comparison with supervised methodologies.

\begin{table}
    \centering
    \caption{Quantitative comparison on $\phi$-SfT synthetic dataset. Average error between ground-truth and predited meshes is reported.}
    \resizebox{1.\linewidth}{!}{%
    \begin{tabular}{@{}rcccc@{}}
    \toprule
    Seq. & Traditional SfT \cite{Bartoli2015} & $\phi$-SfT \cite{Kairanda2022} & PGSfT \cite{Stotko2024} & \emph{Ours} \\
    \midrule
    S1 & 0.0328 &  0.0420 & 0.0298  & \textbf{0.0229}  \\
    S2 & 0.0483 & \textbf{0.0230} & 0.0420 & 0.0254  \\
    S3 & 0.0481 & \textbf{0.0330} & 0.0823 & 0.0357  \\
    S4 & 0.0232 & 0.0050 & 0.0919  & \textbf{0.0031} \\
    \bottomrule
\end{tabular}
}
\label{tab:phi-sft-synthetic} 
\end{table}

\subsection{Evaluation on Real $\phi$-SfT Dataset}
We evaluate our method on 9 real sequences in the $\phi$-SfT dataset and compare our performance with $\phi$-SfT and PGSfT, the best performing state of the art methods. We compute errors as average chamfer distance between ground truth point clouds and predicted point clouds (recovered from sampling on the reconstructed meshes the same number of points as the ground-truth) on each video sequence.  \Cref{tab:quantitative_comparison} shows that our method outperforms $\phi$-SfT \cite{Kairanda2022} and PGSfT \cite{Stotko2024} in all sequences by large margins. Our vision-based adaptive data losses allow us to recover finer details of the object motion and geometry.

$\phi$-SfT \cite{Kairanda2022} is computationally expensive. On average, it processes a single frame in $2800s$. Our runtime performance is comparable to PGSfT \cite{Stotko2024}, which is a self-supervised technique. Our frame-wise optimization strategy, detailed in~\Cref{sec:frame-wise-optimization} 
simplifies optimization scheme used in $\phi$-SfT and PGSfT by large margins. Table \ref{tab:runtime} shows that our runtime in minutes, over $\phi$-SfT real sequences matches PGSfT \cite{Stotko2024}. We refer to Appendix \ref{appendix:complexity-analysis} for more details on complexity comparison of these methods with respect to \textit{Ours}.

\begin{table}
    \centering
    \caption{Quantitative comparison $\phi$-SfT real dataset. Average chamfer distance between ground-truth and predicted points is reported. All values are multiplied by $10^4$ to improve readability.}
    \begin{tabular}{crrr}
        \toprule
        Seq. & $\phi$-SfT~\cite{Kairanda2022} & PGSfT~\cite{Stotko2024} & \emph{Ours} \\
        \midrule
        R1 & 9.36 & 6.05 & \textbf{0.66} \\
        R2 & 11.60 & 4.11 & \textbf{1.30} \\
        R3 & 6.76 & 10.41 & \textbf{4.49} \\
        R4 & 11.59 & 13.19 & \textbf{8.15} \\
        R5 & 14.93 & 11.97 & \textbf{8.04} \\
        R6 & 9.95 & 15.46 & \textbf{3.37}\\
        R7 & 12.34 & 7.20 &  \textbf{4.62} \\
        R8 & 3.23 & 10.21 & \textbf{2.76} \\
        R9 & 3.71 & 9.18 & \textbf{2.12}\\
        \bottomrule
    \end{tabular}
    \label{tab:quantitative_comparison}
\end{table}

\begin{table*}
    \centering
    \caption{Runtime comparison with PGSfT \cite{Stotko2024}. All numbers represent the runtime (in minutes) for the optimization loop until convergence.}
    \setlength{\tabcolsep}{7pt}
    \begin{tabular}{rccccccccc}
        \toprule
         Method & R1 & R2 & R3 & R & R5 & R6 & R7 & R8 & R9 \\
         \midrule
         PGSfT~\cite{Stotko2024} & 2.62 & 2.37 & 2.39 & 2.32 & 2.61 & 2.36 & 2.41 & 2.38 & 2.31 \\
         \emph{Ours} & 3.58 & 2.88 & 2.75 & 2.69 & 2.21 & 2.91 & 2.83 & 2.76 & 2.63 \\
         \bottomrule
    \end{tabular}
    \label{tab:runtime}
\end{table*}

\begin{figure}
    \centering
    \includegraphics[width=1\columnwidth]{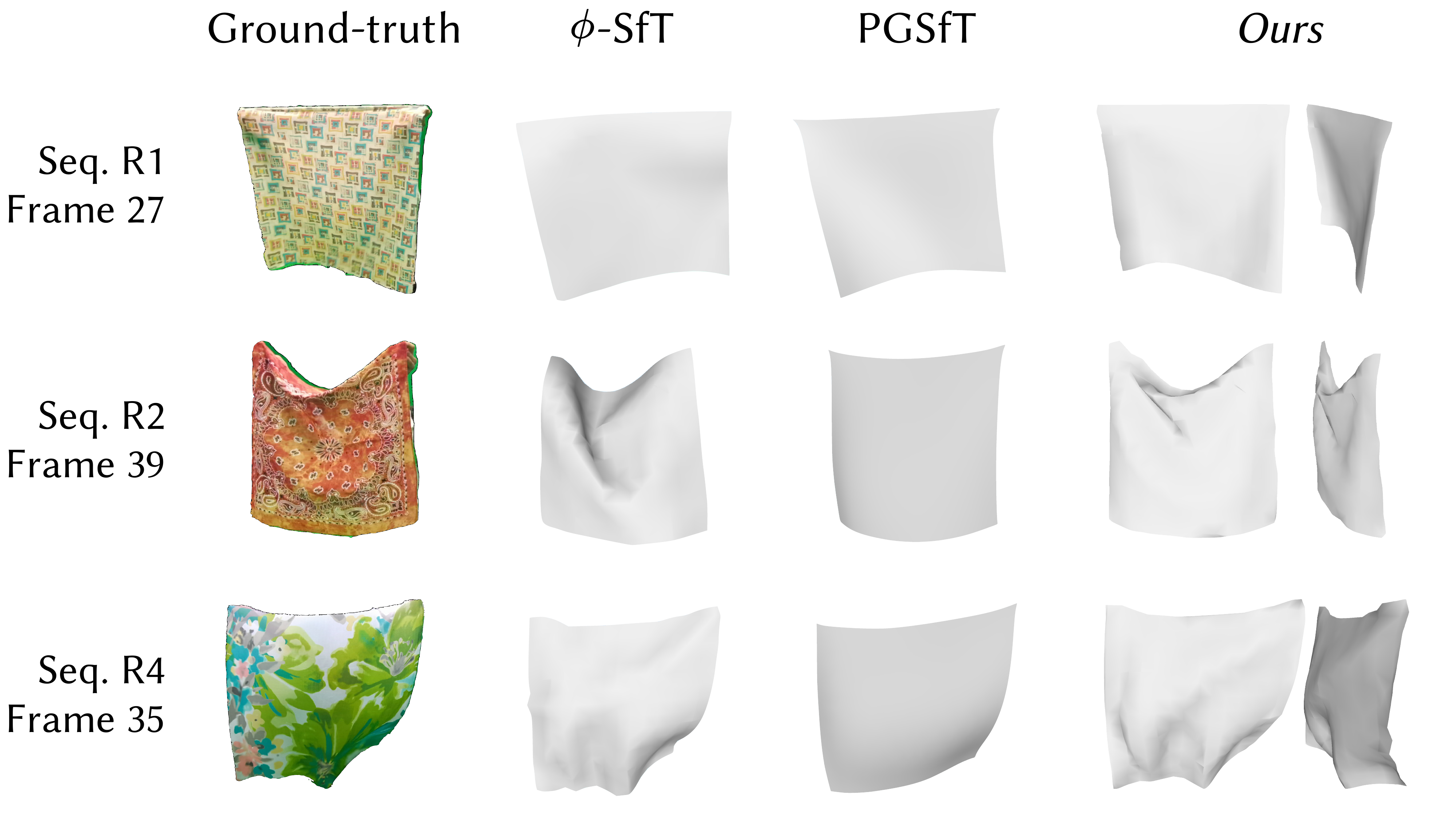}
    \caption{Qualitative comparison on $\phi$-SfT real sequences.}
    \label{fig:qualitative_result}
\end{figure}

\noindent {\textbf{Sharp folds, strong motion and high perspectives.}} Figure \ref{fig:teaser} shows some visual results on a sample frame from R3 and R5 sequence. \emph{Ours} closely match the ground-truth folds and contours, outperforming the other approaches. The traditional SfT\cite{Bartoli2015} underestimates surface variation in certain regions (e.g., the upper corner in R3) due to noises in correspondences computed from CoTracker v3 \cite{karaev2024cotracker3}. The physics-based methods $\phi$-SfT and PGSfT are slightly better but they still fail to capture subtle curvatures accurately.

Figure \ref{fig:qualitative_result} shows the 3D reconstruction of our method from frontal and side perspectives, together with the ones from $\phi$-SfT \cite{Kairanda2022} and PGSfT \cite{Stotko2024}. In sequence R1, the deformation is relatively simple, with a straight upper boundary throughout the frames and only slight movement in the lower region. Hence, our approach achieves an accurate reconstruction with minimal difficulty. Although the rendered frame lacks shading effects, the side view reveals that our method successfully captures the surface folds, which are absent in the other two methods. The sequences R2 and R4 contain high-frequency movement, due to manual folding at the corners and strong wind turbulence, respectively. By using low-resolution templates in $\phi$-SfT dataset \cite{Kairanda2022}, \emph{Ours} captures fine-winkle and high-curvature details very well. Moreover, our reconstructions are smooth and physically plausible without using any smoothness regularization. These detailed reconstructions are attributed to deformation modeling, adaptive loss implementation strategy and efficient frame-wise optimisation, as discussed in Section \ref{sec:method}. Compared to \emph{Ours}, $\phi$-SfT could reconstruct some folds but they are not well aligned with the object shapes. PGSfT, on the other hand, over-smoothed the surfaces and cannot capture any much detail.

\begin{figure}
    \centering
    \includegraphics[width=1.\columnwidth]{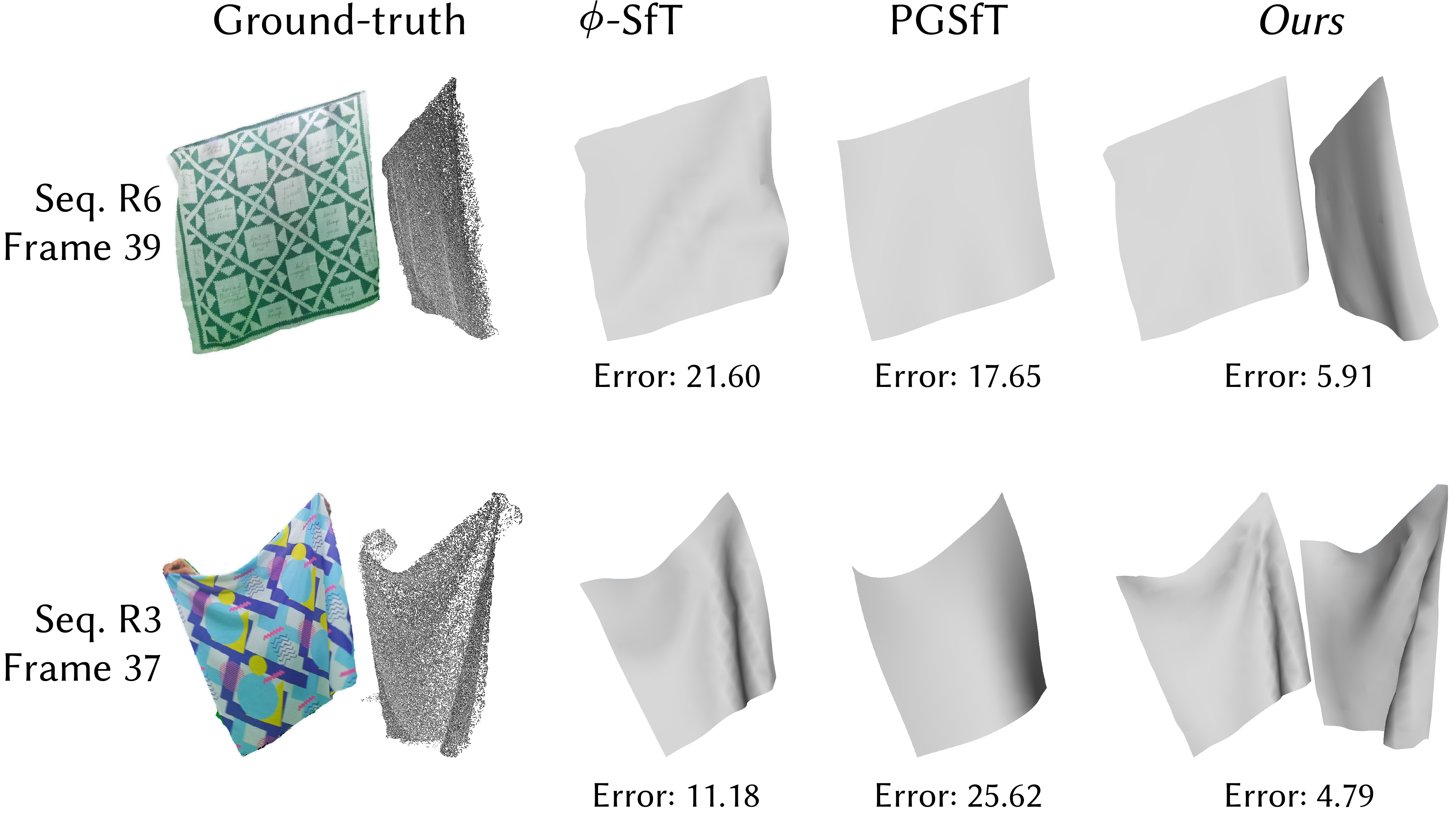}
    \caption{Comparison on frames with strongest self-occlusions on R6 and R3 sequences of the $\phi$-SfT real dataset.} 
    \label{fig:self_occlusion}
\end{figure}

\noindent {\textbf{Self-occlusions.}} Among $\phi$-SfT real sequences, we observe that R3 and R6 have strongest self-occlusion in their final frames. When such behaviors occur, especially over a large number of frames, the surface reconstruction task becomes more challenging for most methods. An image-guided approach typically benefits from temporal constraints to handle the invisible regions.~$\phi$-SfT\cite{Kairanda2022} shows that their physics-based model could provide reasonable results for self-occluded parts. Indeed, their incremental learning has built-in temporal consistency along with Newtonian dynamics. This is also incorporated into the pre-trained neural cloth model proposed by PGSFT \cite{Stotko2024}. However, training such a general-purpose model in self-supervised approach restricts the performance to having low-frequency wrinkles. In contrast, our frame-wise strategy employs no explicit temporal constraint yet still handles self-occlusion effectively. As illustrated in Figure \ref{fig:self_occlusion}, our reconstructed surfaces for both R3 and R6 align more closely to the ground-truth point clouds than the other physics-based methods. We attribute this capability to our initialization scheme in Section \ref{sec:frame-wise-optimization}, in which each frame’s deformation network starts from the suboptimal parameters of the preceding frame.

\subsection{Ablation Study}

\noindent \textbf{Loss Functions.}
\begin{table}
    \centering
    \caption{Ablation study evaluating the impact of removing image gradient and adaptive data loss terms on our performance. Average chamfer distance ($\times 10^4$) on $\phi$-SfT real dataset is reported.}\label{tab:ablation_quantitative}
    \begin{tabular}{cccc}
        \toprule
        Seq.  & No image grad. & No adaptive data loss & Default \\
        \midrule
        R1  & 0.71 & \textbf{0.52} & 0.66 \\
        R2  & \textbf{1.15} & 1.44 & 1.30\\
        R3  & \textbf{4.23} & 9.73 & 4.49\\
        R4  & 8.61  & 11.1 & \textbf{8.15}\\
        R5  & 7.47  & 8.05 & \textbf{8.04}\\
        R6  & 4.09 & 5.85 & \textbf{3.37} \\
        R7  & 5.23 &  \textbf{4.37} & 4.62\\
        R8  & 2.36 & \textbf{2.22} & 2.76\\
        R9  & 1.71 & \textbf{1.56} & 2.12\\
        \midrule
        Avg. & 3.95 & 4.98 & \textbf{3.91} \\
        \bottomrule
    \end{tabular}
\end{table}
The RGB and silhouette loss, together with the mesh inextensibility is quite crucial for every SfT method. We have additionally imposed image gradient loss. In addtion, we have enforced all losses in an adaptive  format. Thus, we evaluate importance of gradients and adaptive data loss in our default setting. We compare three configurations of our loss functions on $\phi$-SfT real sequences \cite{Kairanda2022}, and report error as chamfer distance between ground truth and prediction. As shown in Table \ref{tab:ablation_quantitative}, removing the image gradient loss sometimes yields marginal improvements as in R2 and R3, but generally increases the error on other sequences. Similarly, omitting the adaptive data loss can help certain cases such as R7-R9 but often leads to higher overall error, especially in sequences R3 and R4. In contrast, our default setting, which employs both gradient and adaptive data losses, tends to strike a balanced compromise, achieving better or comparable results in most sequences. We study more about the adaptive data loss, silhouette loss and mesh inextensibility loss in Appendix~\ref{appendix:loss}.

\begin{figure}
    \centering
    \includegraphics[width=1.\columnwidth]{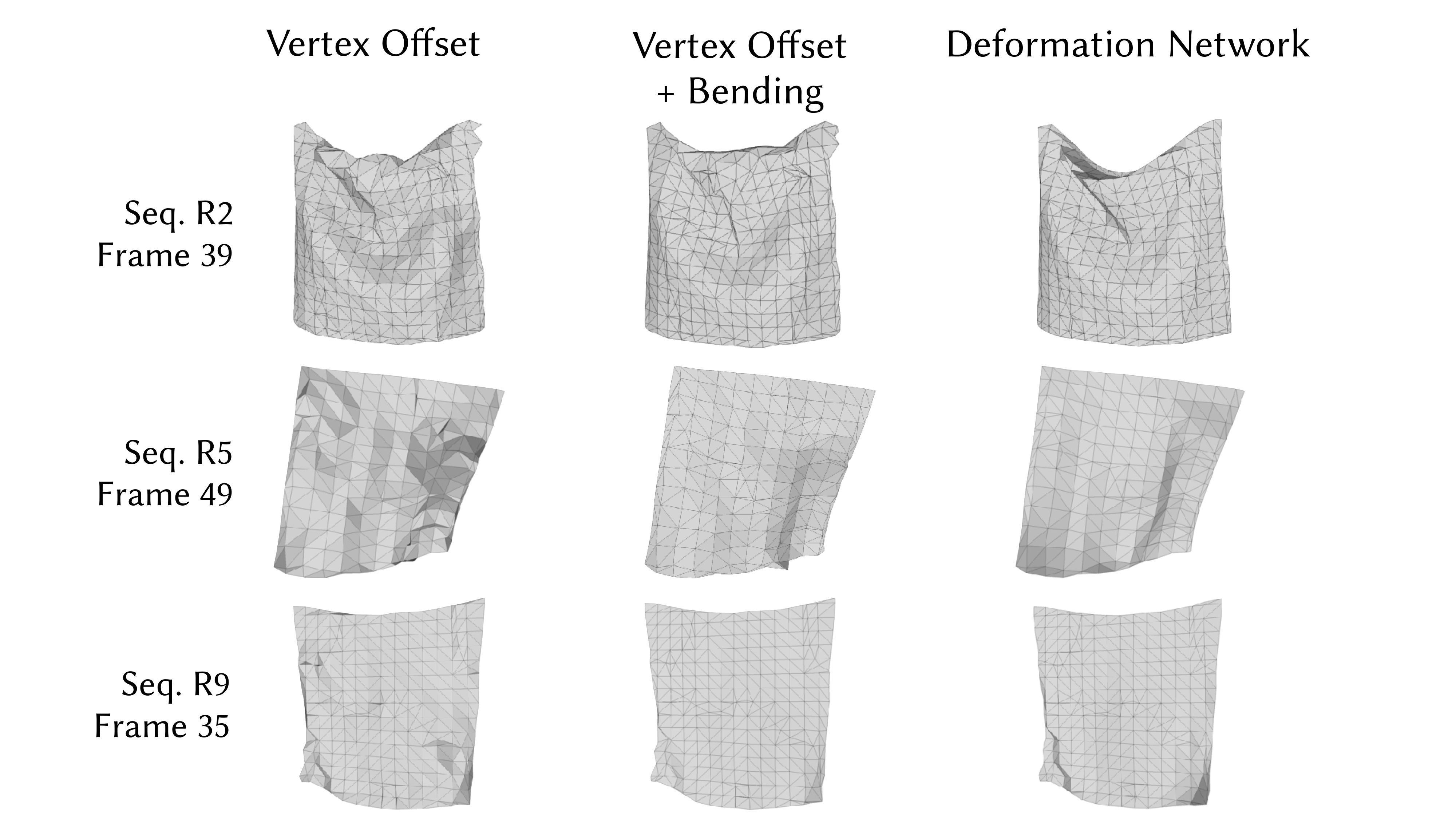}
    \caption{Qualitative comparison on modeling formulation.}
    \label{fig:smoothness}
\end{figure}

\noindent \textbf{Deformation Network.}
To evaluate the deformation model proposed in Section \ref{sec:deformation}, we run our optimization on the 
$\phi$-SfT real sequences under the default configuration, but replacing our deformation network with a vertex-offset parameterization as illustrated in first column of Figure \ref{fig:smoothness}. The resulting meshes contain noticeable non-smoothness compared to the deformation network outputs. We then incorporate the discrete thin-shell bending energy from \cite{sassen2024repulsive} to examine whether this additional regularization can rectify the artifacts. Although it does yield smoother surfaces, the high-frequency folds are lost, indicating that the bending term enforces a trade-off during optimization. As mentioned in~\Cref{sec:deformation}, our deformation network is naturally a continuous latent model that can capture the essential deformations and leave out redundant degrees of freedom that can cause unrealistic shapes.

\begin{figure}
    \centering
    \includegraphics[width=1.\columnwidth]{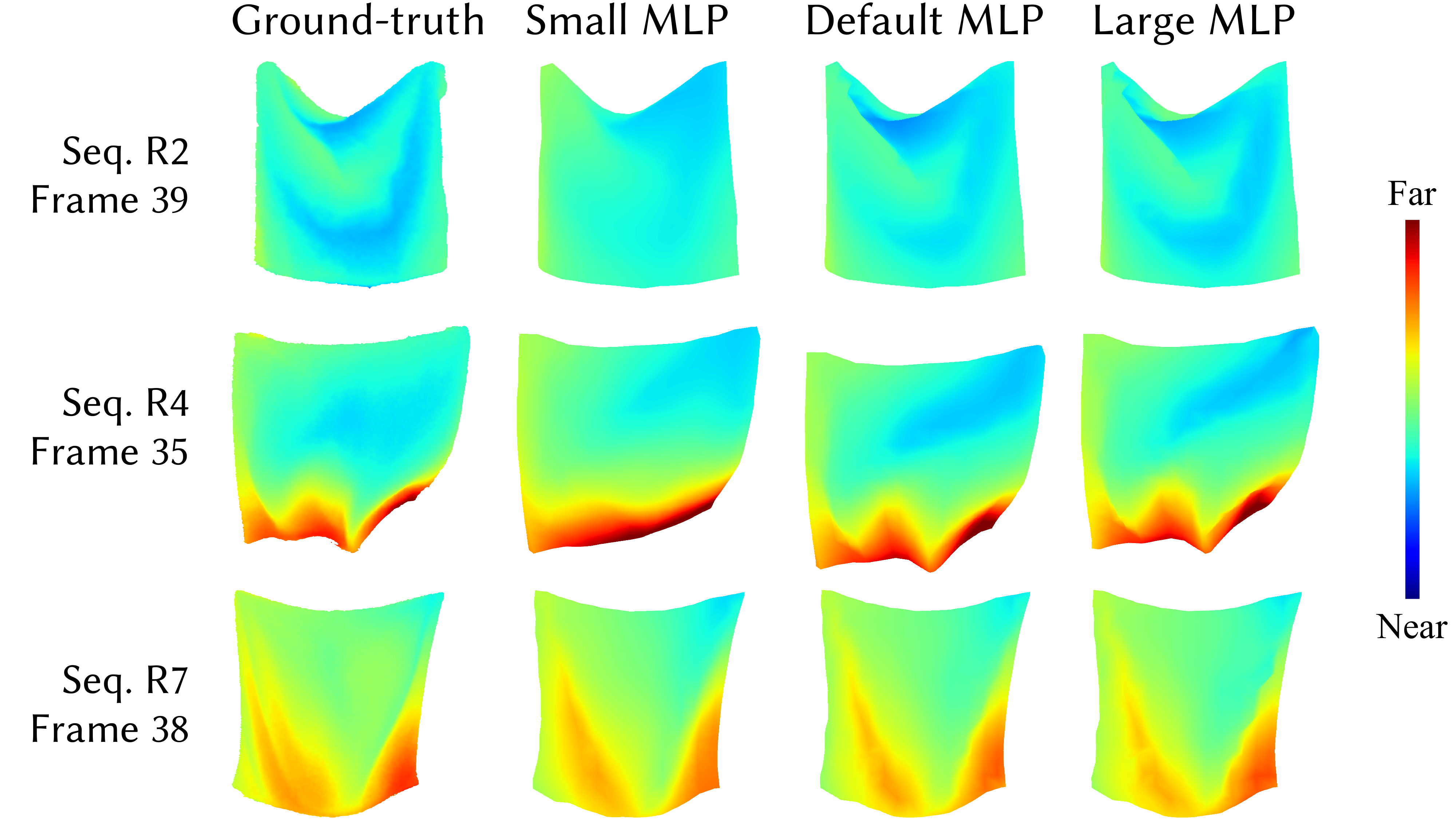}
    \caption{Qualitative comparison on network architecture.}
    \label{fig:ablation_network_architecture}
\end{figure}

\noindent \textbf{Network Architecture.}
In addition to our base ($8$-layer, $256$-width) deformation network, we analyze the network architecture with a small ($4$-layer, $64$-width) and a large ($12$-layer, $512$-width) version. We report the  average reconstruction error in terms of chamfer distance between ground truth and predicted data of the three architectures on $\phi$-SfT real sequences in Table \ref{tab:network_architecture}. We can see that the larger networks generally give better performance due to their greater capacity, which might be able to reach a suboptimal solution more quickly. The small one, on the other hand, struggles to capture high-frequency wrinkles as shown in Figure \ref{fig:ablation_network_architecture} on sequences R2 and R4. For sequence R7, smaller networks perform much better quantitatively, where, in Table \ref{tab:network_architecture}, we have $4.95$ and $4.62$ compared to $6.00$ of the large one.  We visually perceive that the motion of the cloth in this sequence is relatively simpler than others.  Therefore, these observations suggests us to select the network architecture which aligns with the complexity of the object deformation within the video sequence. Moreover, to handle a more complex motion sequence, one can simply increase the network capacity without incurring a computational overhead. Hence, in our experiemnts, we have chosen between base and large MLP, depending on the complexity of the sequence. However, in sequences where large MLP outperforms our base one, the performance gap is not huge. It is important to note that learning on these three networks has almost no difference in runtime.  





\begin{table}
    \centering
    \caption{Quantitative comparison on network architecture. Average chamfer distance ($\times 10^4$) on $\phi$-SfT real dataset is reported.}
    \begin{tabular}{crrr}
        \toprule
        Seq. & Small & Base & Large \\
        \midrule
        R1 & 0.75 & 0.69 & \textbf{0.63} \\
        R2 & 2.49 & 1.30 & \textbf{0.96} \\
        R3 & 4.00 & 4.49 & \textbf{3.11} \\
        R4 & 15.85 & \textbf{8.15} & 8.38 \\
        R5 & 9.97 & \textbf{8.04} & 9.11 \\
        R6 & 4.73 & 3.37 & \textbf{3.02}\\
        R7 & 4.95 & \textbf{4.62} &  6.00 \\
        R8 & 4.06 & 2.76 & \textbf{2.15} \\
        R9 & 3.74 & 2.12 & \textbf{1.73} \\
        \bottomrule
    \end{tabular}
    \label{tab:network_architecture}
\end{table}

\subsection{Limitations}
We noticed that the performance on Kinect Paper dataset is slightly worse than TD-SfT~\cite{Sanchez2025} due to some dips while reconstructing frames with relatively high inter-image motion, as shown at frame 170 in \Cref{fig:kinect_paper}. In these cases, we need more than 200 iterations or other optimization strategies as discussed in Appendix \ref{appendix:optimization-strategy}. Moreover, although the adaptive data loss structure is designed from an intuition about shading, our formulation cannot reconstruct textureless or specular surfaces, which requires for light source and reflectance modeling in shape-from-shading literature \cite{zhang2002shape}.
\section{Conclusion}
In this paper, we presented an unsupervised approach to template-based 3D reconstruction using only vision cues (images, gradients and silhouettes) and mesh inextensibility regularization; thus providing the very first  SfT solely guided by image observations. We introduced an adaptive strategy to enforce vision losses, a function-based deformation network to estimate deformations and an implicitly temporally-coherent frame-wise optimization strategy to devise a simple methodology which not only outperforms state-of-the-art methods by large margins but also achieves a $400\times$ faster computation than the best-performing counterpart, $\phi$-SfT. We also tested several aspects of our proposed methodology and network architectures.
Although our work focuses on triangle meshes, it is worth noting that other representations, such as parametric surfaces, implicit fields, or point-based structures, can also be adapted. We plan to incorporate them in future works. We also plan to extend our frame-wise to a window-wise approach, which allows us to constrain the object shapes with temporal consistency loss
to improve the overall performance in terms of timing and accuracy. Moreover, we would like to discover other object categories that have servere self-occlusion.

\noindent\textbf{Acknowledgments.} This work is supported by the project RHINO (ANR-22-CE33-0014-01), a JCJC research grant. We also thank $\phi$-SfT~\cite{Kairanda2022} and PGSfT~\cite{Stotko2024} authors for releasing the dataset, the code, and their valuable discussions about the benchmarks.
{
    \small
    \bibliographystyle{ieeenat_fullname}
    \bibliography{main}
}
\clearpage
\setcounter{page}{1}
\maketitlesupplementary



\section{Deformation Network}\label{appendix:deformation-network}
The object deformation at time step $t$ can be represented by $\varphi(\mathbf{x}_0,t;\mathbf{z}_t)=\mathbf{x}_t-\mathbf{x}_0$, for the initial state $\mathbf{x}_0$ and latent variables $\mathbf{z}_t$. We can learn such deformation in the following ways:
\begin{enumerate}[label={(\arabic*)}]
    \item \textbf{vertex-offset}: consider $\varphi:=\Delta\mathbf{x}_t\in\mathbb{R}^{3V}$ as learning variables directly,
    \item \textbf{physics-based}: sequentially predict the per-time-step acceleration $\mathbf{a}_t$ and compute the shape using backward Euler method
    \begin{align*}
        \mathbf{v}_{t}&=\mathbf{v}_{t-1}+\Delta t \mathbf{a}_t,\\
        \mathbf{x}_t&=\mathbf{x}_{t - 1} + \Delta t \mathbf{v}_t,
    \end{align*}
    \item \textbf{shared-weight network}: learn a network $\varphi(\mathbf{x}_0,t;\theta)$ with the shared parameters $\theta$ and timestep $t$ as input,
    \item \textbf{frame-wise network} (\textit{ours}): learn a network $\varphi(\mathbf{x}_0;\theta_t)$ with distinct parameters $\theta_t$ for each timestep $t$.
\end{enumerate}
The first idea is intuitive, on which we made an ablation study (see \Cref{fig:smoothness}) to show that direct estimation may lack smoothness and needs additional regularizers.
The state-of-the-art physics-based methods, to which we compare in \Cref{sec:evaluation}, follow the second way. The third approach are widely used in Neural Radiance Fields (NeRF) \cite{mildenhall2021nerf} and Gaussian Splatting (GS) \cite{yang2024deformable}, which should optimize for the entire video concurrently so that the input $t$ is informative. In our initial exploration, we modeled the SfT problem as an initial value problem and utilized neural ODE framework \cite{chen2018neural} to learn the sequence's velocity field. We found it to be as computationally inefficient as $\phi$-SfT. Our method, on the other hand, allows optimizing in a frame-wise strategy described in \Cref{sec:frame-wise-optimization}. This brings the novelty in our work, where we can reconstruct high-quality shapes efficiently. 

\section{Loss Functions}
\label{appendix:loss}
\subsection{Adaptive Data Loss}\label{appendix:adaptive-data-loss}
\begin{figure}
    \centering
    \includegraphics[width=1.0\linewidth]{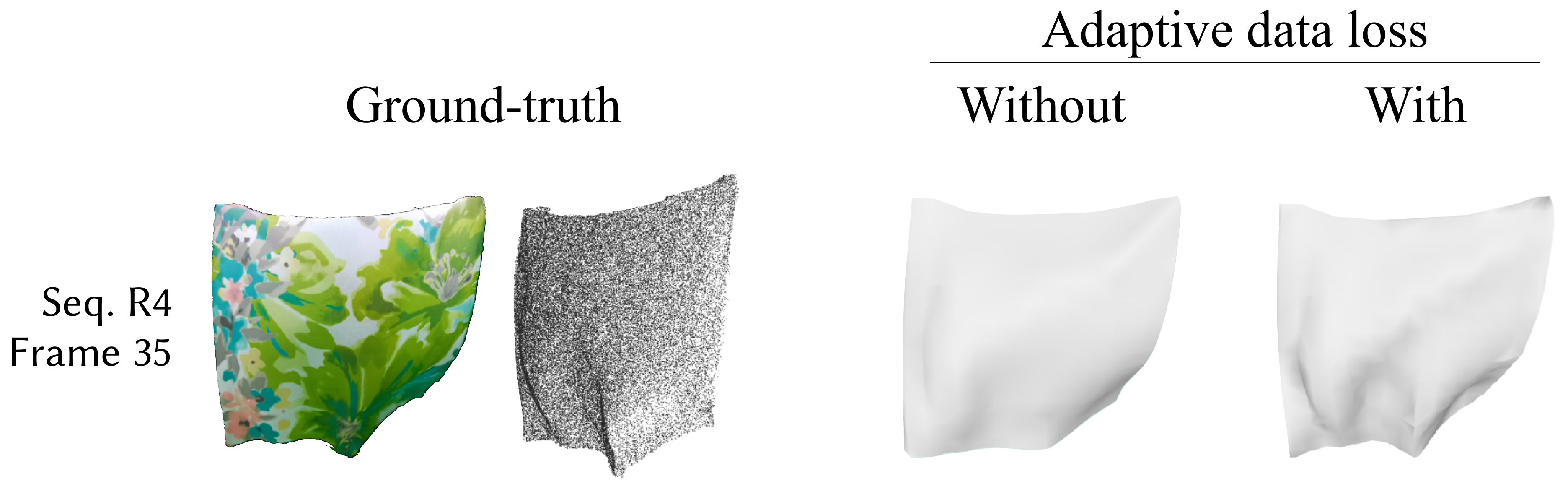}
    \caption{Qualitative comparison on using the adaptive data loss.}
    \label{fig:impact-adaptive-data-loss}
\end{figure}

Rather than using the standard $\ell$-p norm for the vision losses as $\phi$-SfT\cite{Kairanda2022} or PGSfT\cite{Stotko2024}, we propose to use an adaptive data loss as described in \Cref{sec:adaptive-data-loss}. Its adaptive nature comes from the exponential weighting factor on the pixel difference $d$, which allows color errors to remain in focus during optimization despite shading variations. Using this loss is particularly beneficial in complex videos with low-textured areas and/or strong deformations. Indeed, the quantitative results in \Cref{tab:ablation_quantitative} showed a significant improvement on complex sequences while slightly degradation (visually indistinguishable) on the simpler ones. Moreover, as illustrated in \Cref{fig:impact-adaptive-data-loss}, our method is able to reconstruct very high-detailed wrinkles with the adaptive data loss. 

The adaptive data loss is computed with an exponential weighting factor $w(d)=\alpha\exp({d}/{\sigma})$, where we fix $\alpha=10$ and $\sigma=1$. To investigate the robustness of the choice of these hyper-parameters, we evaluate the reconstruction results on $\phi$-SfT real dataset with ranges of $\alpha\in\{0.1,1,5,10,15\}$ and $\sigma\in\{0.1,0.5,1,2,10\}$, then report the average of all sequences in \Cref{table:hyper-param-tuning}. First, we can see that we obtain bad reconstruction when $\sigma=0.1$. Small $\sigma$ makes the weight and thus the adaptive data loss exponentially large which can cause optimization to fail. Next, when $\alpha=0.1$, we also get bad results since the optimizer might lose the importance of vision losses. Finally, when $\alpha$ and $\sigma$ are bigger, the results are quite robust compared to our default setting.

\begin{table}
\centering
\caption{Ablation study on hyper-parameters of the adaptive data loss. Average chamfer distance ($\times 10^4$) are reported in average of all sequences on $\phi$-SfT real dataset.}\label{table:hyper-param-tuning}
$\begin{array}{@{}l*{5}{r}@{}}
   \sigma \text{ }\backslash & \alpha=0.1 & 1 & 5 & 10 & 15\\
\midrule
0.1 & 5.59 & 17.91 & 28.22 & 28.86 & 26.73\\
0.5 & 18.43 & 4.85 & 4.01 & 4.22 & 4.76\\
1 & 18.89 & 4.95 & 3.83 & 3.91 & 4.23\\
2 & 19.91 & 4.75 & 3.93 & 3.95 & 4.18\\
10 & 27.87 & 4.81 & 3.95 & 3.92 & 4.23
\end{array}$
\end{table}

\subsection{Silhouette Loss}\label{appendix:silhouette-loss}
\begin{table}
    \centering
    \caption{Ablation study evaluating the impact of removing silhouette loss terms on our performance. Average chamfer distance ($\times 10^4$) on $\phi$-SfT real dataset is reported.}\label{tab:ablation_silhouette}
    \begin{tabular}{cccc}
        \toprule
        Seq.  & No silhouette loss & Default \\
        \midrule
        R1  & 0.71 & \textbf{0.66} \\
        R2  & \textbf{1.04} & 1.30\\
        R3  & 4.94 & \textbf{4.49}\\
        R4  & \textbf{6.57} & 8.15\\
        R5  & 8.68 & \textbf{8.04}\\
        R6  & 3.56 &\textbf{3.37} \\
        R7  & 4.79 & \textbf{4.62}\\
        R8  & 3.33 & \textbf{2.76}\\
        R9  & 2.18 & \textbf{2.12}\\
        \midrule
        Avg. & 3.97 & \textbf{3.91} \\
        \bottomrule
    \end{tabular}
\end{table}
We inherited the silhouette loss from the related works $\phi$-SfT \cite{Kairanda2022} and PGSfT \cite{Stotko2024}. Hence, the object masks are required in order to optimize the shapes. In case the ground-truth masks are not given, we can estimate them by using a segmentation network, as shown our pipeline (see \Cref{fig:pipeline}). The Kinect paper dataset does not contain object masks, and we used SAM 2 in our experiments. We additionally report in \Cref{tab:ablation_silhouette} the impact of using silhouette loss on $\phi$-SfT real dataset. In general, we can see that the silhouette loss helps improve the reconstruction performance.

\subsection{Mesh Inextensibility Loss}\label{appendix:mesh-inextensibility-loss}
We note that the mesh inextensibility loss is essential in our method. Learning shapes without such a regularization, the object shapes can be overstretched out of viewing frame, thus misguiding the optimization in subsequent iterations. This is similar to using physics constraints as $\phi$-SfT \cite{Kairanda2022} and PGSfT \cite{Stotko2024}, and bypassing the optimization or prior of cloth's material coefficients. Although classical constraints as-rigid-as-possible (ARAP) \cite{Parashar2015} or edge preservation can also be imposed considering their computational efficiency, they are still prone to high errors when we have strong deformations or servere occlusion. As suggested by Chen et al. \cite{Chen2024}, the mesh inextensibility loss relaxes isometry to a weaker form of local rigidity of a point within a neighborhood via preservation of singular values in \Cref{eq:inextensibility-loss,eq:covariance-matrix,eq:template-covariance-matrix}. We refer to their paper \cite{Chen2024} for more discussion on the loss.

The mesh inextensibility loss is originally designed to garment simulation, in which loss variations are not quite large. On the other hand, SfT problem aims to reconstruct various object categories where the inextensibility losses vary a lot. Hence, selecting the weighting factor $w_\text{inext}$ is essential for our framework to generalize to any object of interest. We observe that the mesh inextensibility loss depends on the mesh scale. Indeed, let us define $\delta$ the scale of the object mesh. We imply that the vector $x_j-\bar{x}_i$ also has scale $\delta$. It follows that the covariance matrices in \Cref{eq:covariance-matrix,eq:template-covariance-matrix} have the scale $\delta^2$. Therefore, the mesh inextensibility loss in \Cref{eq:inextensibility-loss} has scale $\delta^6=(\delta^2)^3$ due to the determinant computation of $3\times 3$ matrices. In our experiments, we thus fix an adaptive weighting factor $w_\text{inext}:=\hat{\delta}^{-6}$, where $\hat{\delta}$ approximates the mesh scale by evaluating the median of the edge lengths of the mesh. Since the weighting factors for vision losses are fixed to 1, integrating them into our framework yields an adaptive manner for reconstructing various 3D objects without extra tuning.

\section{Optimization Strategies}\label{appendix:optimization-strategy}
We analyze the complexity of the proposed frame-wise optimization strategy in \Cref{appendix:complexity-analysis} in comparison with the strategies from physics-based approaches. Additionally, we discuss some possible extensions in \Cref{appendix:window-wise,appendix:adaptive-frame-wise}. Depending on the practical applications, future works can develope from these fundamentals to balance between the accuracy and efficiency. We make an ablation study of such strategies on the Kinect Paper dataset in \Cref{appendix:optimization-strategy-ablation-study}, showing promising results on reconstructing high inter-frame motion. 

\subsection{Frame-wise Optimization}
\label{appendix:complexity-analysis}
We compare our proposed optimization strategy in \cref{sec:frame-wise-optimization} with the two state-of-the-art methods $\phi$-SfT \cite{Kairanda2022} and PGSfT \cite{Stotko2024}. A conspicuous bottleneck for image-guided SfT is rendering and backpropagation through the differentiable renderer, especially when the mesh has a high resolution. We show in the following that our strategy reduces the number of renderings from $O(T^2N)$ to $O(TN)$, where $T$ is the video sequence length and $N$ is the average number of iterations used to optimize a single frame. 

Let us first analyze the number of renderings used in $\phi$-SfT \cite{Kairanda2022} and PGSfT \cite{Stotko2024}. These methods rely on the simulation of cloth under Newtonian dynamics. Hence, their proposed optimization strategies are inherently incremental, which requires rendering of prior frames when optimizing the subsequent ones to avoid error accumulation. Assume that the learning process starts with the first frame and gradually incorporates the next frame after $N$ iterations. We can deduce that frame $t$ is rendered $(T-t + 1)N$ times, for $t=1,\dots,T$, which leads to the total number of renderings
\begin{align*}
    \sum_{t=1}^T (T-t+1)N
    &=N\left[\sum_{t=1}^T (T+1)-\sum_{t=1}^T t\right] \\
    &=N\left[T(T+1)-\dfrac{T(T+1)}{2}\right] \\
    &=N\dfrac{T(T+1)}{2}\\
    &=O(T^2N).
\end{align*}
On the other hand, our method renders exactly $TN$ times as each frame is optimized independently. Therefore, our frame-wise strategy scales better for long sequences.

It is to be noted that PGSfT \cite{Stotko2024} leverages physical priors from a pre-trained surrogate model and uses a much smaller number $N$ than $\phi$-SfT \cite{Kairanda2022} and ours, in order to to speed up the optimization process. In future work, this idea can be combined with our strategy on a specific type of object for real-time reconstruction.

\subsection{Window-wise Optimization}\label{appendix:window-wise}
Our strategy for optimizing each frame independently can be extended to optimizing each window of $w$ frames concurrently, where frame-wise strategy is a special case by taking $w=1$. Using more than one frames allows us to employ temporal smoothness regularizer, e.g. for $w=3$, we can impose
\begin{align}
    \mathcal{L}_\text{temporal}(\textbf{x}_t,\textbf{x}_{t+1},\textbf{x}_{t+2})=\ell\left(\textbf{x}_{t+1},\dfrac{\textbf{x}_{t}+\textbf{x}_{t+2}}{2}\right),\label{eq:temporal-loss}
\end{align}
for $\ell$ any choice of data loss. This represents the constraint $\mathbf{x}_{t+1}=\dfrac{\textbf{x}_{t}+\textbf{x}_{t+2}}{2}$ derived from kinematics equations.

\subsection{Adaptive Frame-wise Optimization}\label{appendix:adaptive-frame-wise}
In our default setting, we used a warmup of $500$ iterations and then $200$ iterations for each frame to reconstruct the shapes within a video. This setting well balances between accuracy and runtime, as illustrated in our experiments in \Cref{sec:evaluation}. However, this fixed-iteration strategy might fail to capture inter-frame complex and fast deformations as seen at frame 170 in \Cref{fig:kinect_paper} on Kinect Paper dataset. Using more iterations can improve the performance, but the number of iterations vary among sequences and need to be tuned appropriately. Therefore, adaptive stopping criteria for per-frame optimization can be developed. A straightforward approach is to fix a tolerance on the difference of the total loss values between two consecutive iterations. 

\subsection{Ablation Study}\label{appendix:optimization-strategy-ablation-study}
\begin{table}
    \centering
    \caption{Ablation study of various optimization strategies on Kinect Paper dataset. Average depth RMSE and runtime are reported.}
    \noindent\resizebox{\linewidth}{!}{
    \begin{tabular}{rrrrr}
        \toprule
         & Default & 250-iter & Window & Adaptive \\
        \midrule
        RMSE (mm) & 4.01 & 3.99 & \textbf{3.57} & 3.64 \\
        Runtime (mins) & \textbf{6.04} & 7.52 & 50.54 & 84.37 \\
        \bottomrule
    \end{tabular}
    }
    \label{tab:ablation_study_kinect_paper}
\end{table}
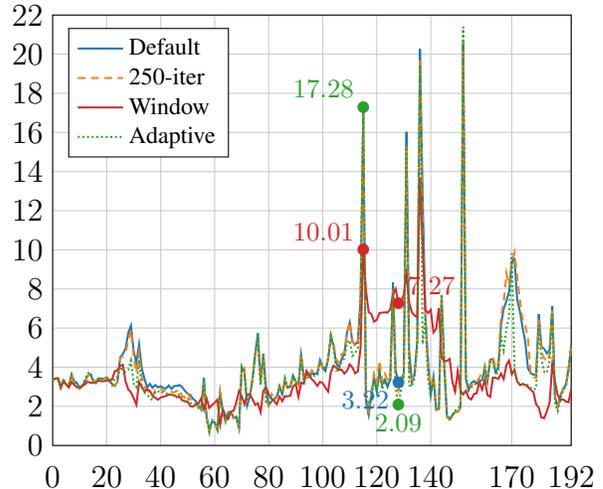
\begin{figure}
    \centering
    \noindent\resizebox{\linewidth}{!}{
\pgfplotsset{xtick style={draw=none}, xtickmin=0, xtickmax=200, xtick align=outside}



\begin{tikzpicture}
\begin{axis}[
  xmin={0}, xmax={192}, domain={1:192}, xmode={linear}, 
  xtick={0, 20, 40, 60, 80, 100, 120, 140, 170, 192}, 
  ymin={0}, ymax={22}, restrict y to domain*=0:22, 
  ytick={0,2,4,6,8,10,12,14,16,18,20,22}, 
  grid={major}, legend pos={north west}, legend cell align={left}, legend style={nodes={scale=0.9, transform shape}, font=\normalsize, /tikz/every even column/.append style={column sep=1.6mm}},
  label style={font=\large},
        tick label style={font=\large} 
]

\addplot[C0, line width=0.75pt] coordinates {\drawDefault};
\addplot[densely dashed, C1, line width=0.75pt] coordinates {\drawMoreIterations};
\addplot[C3, line width=0.75pt] coordinates {\drawWindowWise};
\addplot[densely dotted, C2, line width=0.75pt] coordinates {\drawAdaptiveFrameWise};



\addplot[C0, only marks, forget plot]  coordinates {(128, 3.2263996601104736)};\node at (axis cs:(128, 3.2263996601104736) [anchor={north east}] {\textcolor{C0}{$3.22$}};

\addplot[C3, only marks, forget plot]  coordinates {(128, 7.27396297454834)};\node at (axis cs:(128, 7.27396297454834) [anchor={south west}] {\textcolor{C3}{$7.27$}};
\addplot[C3, only marks, forget plot]  coordinates {(115, 10.019689559936523)};\node at (axis cs:(115, 10.019689559936523) [anchor={south east}] {\textcolor{C3}{$10.01$}};

\addplot[C2, only marks, forget plot]  coordinates {(128, 2.091752052307129)};\node at (axis cs:(128, 2.091752052307129) [anchor={north}] {\textcolor{C2}{$2.09$}};
\addplot[C2, only marks, forget plot]  coordinates {(115, 17.28974723815918)};\node at (axis cs:(115, 17.28974723815918) [anchor={south east}] {\textcolor{C2}{$17.28$}};

\legend{
  {Default},
  {250-iter},
  {Window},
  {Adaptive}
}
\end{axis}
\end{tikzpicture}
    }
    \caption{Illustration of depth map RMSE (mm) on the entire Kinect Paper sequence of 193 frames with various optimization strategies.}
    \label{fig:ablation_study_kinect_paper}
\end{figure}
We experiment on Kinect Paper dataset using the following optimization strategies:
\begin{enumerate}[label={(\arabic*)}]
    \item \textbf{Default}: the default frame-wise strategy with $500$-iteration warmup and $200$ iterations per frame,
    \item \textbf{250-iter}: the frame-wise strategy with $500$-iteration warmup and $250$ iterations per frame,
    \item \textbf{Window}: the window-wise strategy in \Cref{appendix:window-wise} $500$-iteration warmup and 600 iterations for each window of $w=3$ frames. We also add the temporal loss in \Cref{eq:temporal-loss} for $\ell(\textbf{x},\textbf{y})=||\textbf{x}-\textbf{y}||_2$.
    \item \textbf{Adaptive}: the adaptive frame-wise strategy in \Cref{appendix:adaptive-frame-wise} with tolerance $\tau=10^{-6}$.
\end{enumerate}
\Cref{tab:ablation_study_kinect_paper} shows the depth map root mean square error (RMSE) and runtime of the four strategies. We additionally plot the per-frame RMSE for the entire sequence in \Cref{fig:ablation_study_kinect_paper}. We can observe that later frames $110$-$170$ are tricky to our method. Using more iterations, \textcolor{C1}{\textbf{250-iter}} improves the accuracy a bit compared to \textcolor{C0}{\textbf{Default}} and thus needs a minute more. However, the overall behavior looks quite similar to \textcolor{C0}{\textbf{Default}}. \textcolor{C3}{\textbf{Window}} can enhance the average accuracy in a temporal smoothness manner on the error curve. This strategy, however, needs more studies on tuning the per-window iterations or window size to achieve better performance. \textcolor{C2}{\textbf{Adaptive}} strategy illustrates its effectiveness of reaching highly accurate reconstruction (e.g. RMSE reduced from \textcolor{C0}{$3.22$} with \textcolor{C0}{\textbf{Default}} and \textcolor{C3}{$7.27$} with \textcolor{C3}{\textbf{Window}}, to \textcolor{C2}{$2.09$} at frame $128$) while still strugles with some tricky frames (e.g. RMSE at frame $115$ is \textcolor{C2}{$17.28$} compared to \textcolor{C3}{$10.01$} with \textcolor{C3}{\textbf{Window}}). Further works can study on the tolerance to improve accuracy, while noticing the costly tradeoff in runtime.

\end{document}